\documentclass[runningheads]{llncs}
\usepackage[T1]{fontenc}
\usepackage{float} 
\usepackage[most]{tcolorbox} 
\usepackage{colortbl} 
\usepackage{enumitem} 
\definecolor{lightgrey}{gray}{0.87}
\newtcolorbox{examplebox}{
  colback=gray!10,   
  colframe=gray!80,  
  boxrule=0.8pt,     
  arc=4pt,           
  left=6pt,          
  right=6pt,         
  top=0pt,           
  bottom=0pt,        
}
\usepackage{amssymb,amsmath,array}
\usepackage{subcaption}
\usepackage{url}
\usepackage{graphicx}
\usepackage{tabularx,tabulary,booktabs}
\usepackage{multirow}

\newcolumntype{b}{X}
\newcolumntype{s}{>{\hsize=.24\hsize}X}
\newcolumntype{g}{>{\hsize=.6\hsize}X}
\begin{document}
\title{Source Attribution in Retrieval-Augmented Generation}
%
\author{Ikhtiyor Nematov\inst{1,2} \and Tarik Kalai\inst{1} \and
Elizaveta Kuzmenko\inst{1} \and
Gabriele Fugagnoli\inst{1,3} \and
Dimitris Sacharidis\inst{1} \and
Katja Hose\inst{4}\and Tomer Sagi\inst{2}}
\authorrunning{Nematov et al.}

\institute{Université Libre de Bruxelles, Belgium \and
Aalborg University, Denmark\\
\and
University of Padova, Italy\\
\and
TU Wien, Austria\\
}
\maketitle              
\begin{abstract}
While attribution methods, such as Shapley values, are widely used to explain the importance of features or training data in traditional machine learning, their application to Large Language Models (LLMs), particularly within Retrieval-Augmented Generation (RAG) systems, is nascent and challenging. The primary obstacle is the substantial computational cost, where each utility function evaluation involves an expensive LLM call, resulting in direct monetary and time expenses. This paper investigates the feasibility and effectiveness of adapting Shapley-based attribution to identify influential retrieved documents in RAG. We compare Shapley with more computationally tractable approximations and some existing attribution methods for LLM. Our work aims to: (1) systematically apply established attribution principles to the RAG document-level setting; (2) quantify how well SHAP approximations can mirror exact attributions while minimizing costly LLM interactions; and (3) evaluate their practical explainability in identifying critical documents, especially under complex inter-document relationships such as redundancy, complementarity, and synergy. This study seeks to bridge the gap between powerful attribution techniques and the practical constraints of LLM-based RAG systems, offering insights into achieving reliable and affordable RAG explainability.

\end{abstract}
\section{Introduction}
Large Language Models (LLMs) have demonstrated remarkable capabilities across a wide range of natural language tasks. Retrieval-Augmented Generation (RAG) has emerged as a powerful paradigm to further enhance Large Language Models (LLMs) by grounding their responses in external knowledge. In RAG systems, relevant documents are first retrieved from a corpus based on an input query, and then these documents are provided as context to an LLM to generate a more informed and factual response. While RAG significantly improves response quality and reduces hallucination, the process often remains a black box: understanding which of the retrieved documents were truly influential in shaping the LLM's final output is a critical yet underexplored challenge.

The ability to attribute an LLM's response in a RAG system back to specific source documents holds immense value. It can enhance user trust by providing transparency, aid in debugging by identifying irrelevant or misleading retrieved documents, and inform the optimization of the retrieval component itself. However, applying traditional feature attribution methods from machine learning to the RAG setting presents unique obstacles. Foremost among these is the computational expense: many attribution techniques require multiple model evaluations with perturbed inputs. In the context of LLMs, each such evaluation involves a call to the model, which can be computationally intensive and, for API-based models, incur direct monetary costs. This high cost of probing the LLM makes exhaustive attribution methods often impractical.

This paper investigates the application of Shapley values \cite{shapley1953}---a principled, game-theoretic approach for fair credit assignment---to the problem of document importance attribution in RAG systems. We define the \textit{utility} of a given subset of retrieved documents, $S$, as the LLM's capability to generate its original, full-context response ($R_{\mathrm{target}}$) when conditioned on that subset $S$ and the query $Q$. This utility is specifically measured by the \textit{log-likelihood} of $R_{\mathrm{target}}$ --- a metric representing how probable the LLM finds that specific word sequence, calculated as the sum of log-probabilities of each word given its predecessors and the context $(Q,S)$. A higher log-likelihood for a subset $S$ indicates its stronger support for generating $R_{\mathrm{target}}$. 
Although Shapley values are grounded in a set of desirable axioms—efficiency, symmetry, dummy, and additivity—these assumptions are generally not satisfied in our RAG setting. The log-likelihood utility is non-additive, and the contributions of documents often depend on their interactions, leading to violations of theoretical guarantees. Nonetheless, Shapley values remain valuable in practice: their averaging over all subsets helps them robustly capture the influence of documents, even in the presence of nonlinearity \cite{datashap}. Our goal is to evaluate whether Shapley-based attribution can still provide meaningful insights in the RAG scenario, and how well it performs compared to computationally cheaper alternatives. We also study how these methods perform in scenarios with intricate inter-document dependencies common in RAG.

Our observations indicate that utility-based attribution methods hold promise for RAG. However, complex inter-relationships among retrieved documents—such as redundancy, complementarity, and synergy—pose significant challenges for the accuracy and interpretability of the resulting attributions.

\section{Foundational Concepts of Attribution}
\label{sec:attribution_in_ml}
Attribution, in the context of complex systems, refers to the process of assigning credit or blame for an outcome to its constituent components or inputs. The goal is to understand how individual parts contribute to a collective result, thereby providing transparency and enabling informed decision-making. At the heart of many attribution methods lies the concept of \textit{utility}, which is a function that quantifies the value, output, or payoff generated by a coalition of players or a subset of components. Given a set of players, attribution methods aim to distribute the total utility achieved by the grand coalition (all players) among individual players according to their contributions.

A cornerstone of attribution is the concept of \textbf{Shapley values}, originating from cooperative game theory~\cite{shapley1953}.
For a set of players $D$ and a characteristic utility function $v(S)$ that defines the worth of any subset (coalition) $S \subseteq D$, the Shapley Value $\phi_j(v)$ of a player $j$ quantifies its average marginal contribution to all possible coalitions:
 \begin{equation}
\phi_j(v) = \sum_{S \subseteq D \setminus \{j\}} \frac{|S|! (|D| - |S| - 1)!}{|D|!} [v(S \cup \{j\}) - v(S)]
\label{eq:shapley_value}
\end{equation}

Shapley values uniquely satisfy four key axioms: \textit{efficiency} (the total utility is fully distributed), \textit{symmetry} (equally contributing features get equal scores), \textit{dummy} (features with no marginal impact get zero credit), and \textit{additivity} (Shapley values for combined games are additive).

Another intuitive and simpler attribution approach is the \textbf{Leave-One-Out (LOO)}. For a player $j$, its contribution is measured by the change in utility when it is removed from the grand coalition: $v(D)-v(D \setminus \{j\})$. While easy to compute and understand, LOO only considers the marginal contribution in the context of all other players being present.

\subsection{Attribution in Machine Learning}
The foundational concepts of attribution have been widely adapted to explain the behavior of machine learning models. We can broadly categorize these methods as follows:

\subsubsection{Utility-based Methods}
These methods aim to define a utility function $v(S)$ related to the model performance for a subset $S$ e.g. loss or other performance metric. In this context, ``players'' can be input features \cite{kernel-shap}, training data instances \cite{datashap,datamodels}.

In \cite{kernel-shap}, the authors introduce \textbf{Kernel SHAP}, a method for feature attribution in machine learning models based on Shapley values. The utility function \( v(S) \) is defined as the model output when the features in \( S \) are retained and the remaining features are masked or replaced with a baseline value. However, this utility function is inherently non-linear and does not generally satisfy the Shapley axioms, such as \emph{additivity} (i.e., if \( S = S_1 \cup S_2 \) and \( S_1 \cap S_2 = \emptyset \), then \( v(S) = v(S_1) + v(S_2) \)). To mitigate this, Kernel SHAP fits a weighted linear regression model that approximates the utility scores for various feature coalitions. It samples coalitions \( S \), estimates \( v(S) \) by evaluating the original model on masked inputs, and fits a linear model to these samples weighted by the SHAP kernel \( \pi_x(S') \), where \( S' \) is the binary indicator vector of the coalition. The sampling procedure favors smaller and larger subsets to ensure informative weights. The resulting feature attribution scores correspond to the coefficients of this surrogate linear model, which satisfy the additivity axiom by construction. 

In a similar manner, Datamodels \cite{datamodels} compute the attribution scores for training data. In this case using a surrogate model not only satisfies the axioms but also makes it more efficient, and avoids retraining the model $2^{|D|}$ times.

\textbf{Truncated-Monte-Carlo-Shapley (TMC-Shapley)} estimates Shapley values by sampling permutations of training data. For a training point $j$, its marginal contribution is calculated based on its predecessors in a sampled permutation $\pi$: $v(P_j^{\pi} \cup \{j\}) - v(P_j^{\pi})$, where $P_j^{\pi}$ are points preceding $j$ in $\pi$. These contributions are averaged over many permutations $T$. Truncation is applied to limit the number of evaluations per permutation. \textbf{Beta Shapley}~\cite{betashap} aims to improve sampling efficiency by recognizing that coalitions of very small or very large sizes often contribute more to the variance of Shapley value estimates. It samples coalition sizes $|S|$ from a Beta distribution, $Beta(\alpha, \beta)$, and then samples coalitions $S$ of that size to estimate marginal contributions. Both Beta Shapley and TMC-Shapley demonstrate that Shapley values can still be effective even when the additivity axiom is violated.

\subsubsection{Model Internals-based Methods}
These methods leverage the internal architecture and parameters of the model, rather than treating it as a black box that only evaluates a utility function.

\paragraph{Gradient-based Methods:} These methods use gradients to measure how the model’s output or loss changes with small input or parameter variations. Simple forms, like saliency maps~\cite{Simonyan2013Deep}, visualize raw gradients to show which input regions affect the output most. Extensions such as SmoothGrad~\cite{Smilkov2017SmoothGrad} average gradients over noisy input copies for clearer maps. Integrated Gradients~\cite{Sundararajan2017Axiomatic} improve interpretability by accumulating gradients along a path from a baseline to the input, ensuring attributions sum to the prediction difference. Influence Functions~\cite{if,basu2021influence} also rely on gradients to estimate how removing or changing a training point would affect the model without retraining. All these approaches need access to internal gradients, so they apply in white-box settings.

\paragraph{Attention-based Methods:} For models with attention, like Transformers~\cite{Vaswani2017Attention}, these methods interpret learned attention weights as indicators of input importance. Attention scores show how much focus the model gives to different input parts when generating an output. To attribute importance to a document, one can aggregate attention from output tokens back to input tokens linked to that document. This requires access to internal attention matrices. However, there is ongoing debate about whether attention weights reliably explain model behavior or true causal influence~\cite{jain-wallace-2019-attention,wiegreffe-pinter-2019-attention}.

\subsection{Attribution in LLM}
\subsubsection{Token-level Attribution}
Many general ML attribution methods have been adapted for LLMs at the token level \cite{sarti-etal-2023-inseq}. Gradient-based methods compute saliency scores for input tokens by backpropagating from an output (e.g., logit of a target token) \cite{contrastllm}. Attention-based methods directly use attention scores within the transformer layers \cite{seat,wra}. 

\subsubsection{Attribution for RAG (Document-level Attribution)}
While proponents of token-level methods suggest that scores can be aggregated (e.g., by summing or averaging) to derive document-level importance, this aggregation step itself can be non-trivial, and the underlying need for model internals (attention weights or gradients) remains a significant barrier for black-box RAG systems.

In RAG each document is a separate unit and attribution must be w.r.t. the whole response. So the utility function $v(S)$ for a document subset $S \subseteq D$ could be defined as the sum of the log-probabilities of the tokens in $R_{\mathrm{target}}$ when conditioned on $Q$ and $S$, effectively measuring $\log P(R_{\mathrm{target}} | Q, S)$. This probability is typically computed using a ``teacher-forcing'' approach: at each step $t$ of generating $R_{\mathrm{target}}$, the model is conditioned on the ground-truth previous tokens $\text{token}_{<t}(R_{\mathrm{target}})$ from the target response itself, rather than its own previously predicted tokens. This allows for a decomposed calculation of the joint probability as a product of individual token probabilities:
\begin{equation}
v(S) = \sum_{t=1}^{|R_{\mathrm{target}}|} \log P(\text{token}_t(R_{\mathrm{target}}) | \text{token}_{<t}(R_{\mathrm{target}}), Q, S)
\label{eq:utility}
\end{equation}
where $token_t$ is the $t^{th}$ token of $R_{target}$.

With this utility function at hand, all utility-based attribution methods become applicable in the RAG setting. This direction was initially explored by \textbf{ContextCite} \cite{contextcite}, which, inspired by methods such as Datamodels \cite{datamodels} and Kernel SHAP \cite{kernel-shap}, trains a simple linear model to predict utility scores and interprets the learned weights as document attribution scores.

In this work, we extend this line of research by systematically investigating how various utility-based attribution techniques can be adapted for document-level attribution in the RAG context.


\section{Evaluation Framework}
\label{sec:evaluation_framework}

This section formalizes the attribution problem for RAG and proposes a comprehensive evaluation protocol designed to assess the effectiveness of various attribution methods.

\subsection{Problem Formalization and Notation}
\label{subsec:problem_definition}

Let $Q$ be an input query. A RAG system first retrieves a set of documents $D = \{d_1, d_2, \ldots, d_n\}$, deemed relevant to $Q$. Subsequently, a LLM generates a response $R_{\mathrm{target}}$ conditioned on both $Q$ and $D$.
The core of our attribution task is to assign an importance score $\phi_i$ to each document $d_i \in D$, reflecting its contribution to the generation of $R_{\mathrm{target}}$.

As in ContextCite \cite{contextcite}, the utility function, $v(S)$, for a subset of documents $S \subseteq D$, is defined as the likelihood of the LLM generating the \textit{initial} response $R_{\mathrm{target}}$ when provided with the query $Q$ and the subset $S$, see Equation \ref{eq:utility}.

An attribution method $\mathcal{M}$ takes $Q$, $D$, and $R_{\mathrm{target}}$ as input and outputs a vector of importance scores $\Phi_{\mathcal{M}} = \{\phi_{\mathcal{M},1}, \phi_{\mathcal{M},2}, \ldots, \phi_{\mathcal{M},n}\}$ for documents in $D$.
\subsection{Attribution Methods}
We adapt the following methods for RAG, where utilities are calculated using the Equation \ref{eq:utility}:
\textbf{Shapley} \cite{shapley1953},
\textbf{TMC-Shapley} \cite{datashap},
\textbf{Beta-Shapley} \cite{betashap},
\textbf{Kernel-SHAP} \cite{kernel-shap},
\textbf{ContextCite} \cite{contextcite}, and
\textbf{LOO}.

Apart from LOO and Shapley, all methods need some ground truth utilities to be calculated for approximation e.g. training a linear model. We experiment with three different numbers ${32, 64, 100}$.
\subsection{Models}
We evaluate all methods using instruction-finetuned versions of three models: \textbf{LLaMA-3.2-8B-Instruct}, \textbf{Mistral-7B-Instruct}, and \textbf{Qwen-3B-Instruct}. Before performing attribution analysis, we first assess the response quality of these models with respect to the ground-truth answers provided in the datasets. Evaluation details can be found in the Appendix.
The codebase for all experiments, method implementations, and datasets is available in our GitHub \footnote{\url{https://github.com/iNema9590/LLMX}}.

\subsection{Evaluation Protocol}

Our evaluation protocol is designed to address three key research questions:
\begin{itemize}
    \item \textbf{RQ1.} How accurately can computationally less exhaustive methods replicate Shapley in the RAG setup?
    \item \textbf{RQ2.} How effective are these methods in identifying the $k$ documents whose removal causes the highest utility drop?
    \item \textbf{RQ3.} How do these methods perform under challenging scenarios involving inter-document relationships that are common in RAG?
\end{itemize}

\subsubsection{Experiment 1: Shapley replication quality}

\paragraph{Procedure}
For each query-document set $(Q, D)$ and its corresponding $R_{\mathrm{target}}$:
\begin{enumerate}
    \item Compute Shapley scores $\Phi_{\mathrm{Shapley}}$. This serves as our reference for the quality of approximation.
    \item For each other method $\mathcal{M}$, compute its attribution scores $\Phi_{\mathcal{M}}$.
\end{enumerate}

\paragraph{Metrics}
\begin{itemize}
    \item \textbf{Pearson Rank Correlation ($\rho$)}: Measures the linear correlation between the ranks of document importances assigned by $\Phi_{\mathcal{M}}$ and $\Phi_{\mathrm{Shapley}}$.
    \item \textbf{Kendall's Tau ($\tau$)}: Measures the ordinal association between the rankings produced by $\Phi_{\mathcal{M}}$ and $\Phi_{\mathrm{Shapley}}$.
    \item \textbf{Precision@k ($P@k_{\mathrm{shap}}$)}: Measures the proportion of the top-$k$ documents identified by $\mathcal{M}$ that are also among the top-$k$ documents identified by Shapley. We will evaluate for $k \in \{1, 3, 5\}$.
\end{itemize}

\paragraph{Datasets} The experiments conducted in two commonly used for RAG datasets: \textbf{BioAsq} \cite{bioasq}, and \textbf{Natural Questions (NQ)} \cite{nq}. From each dataset, we choose 100 queries, each with 10 corresponding documents, 5 of which are \emph{relevant}, i.e., can contribute to the answer, 4 are \emph{hard negatives}, i.e., same topic but don't contribute to the answer, and 1 \emph{soft negative}, i.e., irrelevant document. We report metrics averaged over these 100 queries.

\subsubsection{Experiment 2: Attribution effectiveness}

\paragraph{Procedure}
For each query-document set $(Q, D)$ and its corresponding $R_{\mathrm{target}}$:
\begin{enumerate}
    \item For a given $k$, we exhaustively identify the subset of $k$ documents $S^*_k \subset D$ (where $|S^*_k| = k$) whose removal results in the maximum decrease in the utility function. That is, 
    \begin{equation}
    S^*_k = \operatorname*{argmax}_{S \subset D, |S|=k} \left( v(D) - v(D \setminus S) \right).
    \label{eq:ground_truth_impact}
    \end{equation}
    The set $D^*_k$ consists of these $k$ most impactful documents. This requires evaluating $\binom{n}{k}$ subsets for removal.
    \item For each attribution method $\mathcal{M}$, obtain its top-$k$ predicted documents, $D_{\mathcal{M},k}$, based on its scores $\Phi_{\mathcal{M}}$.
    \item Compare $D_{\mathcal{M},k}$ with $D^*_k$.
\end{enumerate}

\paragraph{Metrics}
\begin{itemize}
    \item \textbf{Precision@k ($P@k_{\mathrm{impact}}$)}: Calculated as $|D_{\mathcal{M},k} \cap D^*_k| / k$. This measures how many of the method's top-$k$ identified documents are actually among the $k$ most impactful documents. We will evaluate for $k \in \{2, 3, 4, 5\}$.
\end{itemize}
\paragraph{Datasets} The same datasets from the previous experiment are used.

\subsubsection{Experiment 3: Inter-document relationships}

\paragraph{Datasets}
We created a synthetic dataset consisting of questions and documents about fictional scenarios to ensure that, when answering the questions, the LLM relies on the provided documents rather than its prior knowledge.

\begin{itemize}
    \item \textbf{Redundancy:} Each query $Q$ is associated with 10 documents $D$ containing one pair of documents $(d_i, d_{i'})$ that are semantically very similar (near-duplicates) and both contain information relevant to answering $Q$.
    \begin{examplebox}
    \textit{\textbf{Question}: 	What is the weather in Suvsambil?} \\
    \textit{\textbf{Documents}:}
    \begin{enumerate}
        \item \textit{The weather in Suvsambil is sunny}
        \item \textit{Suvsambil is a mountainous country}
        \item \textit{The sun is shining in Suvsambil today}
    \end{enumerate}
    
    The LLM can answer the question using either first or third document.
    \end{examplebox}
    \item \textbf{Complementarity:} Each query $Q$ requires information from two distinct documents $d_i, d_j \in D$ to be answered comprehensively. Each such document provides a unique, non-overlapping piece of the answer.
    \begin{examplebox}
    \textit{\textbf{Question}: 	What are the roles or professions of Elara Vayne and Jax Korden?} \\
    \textit{\textbf{Documents}:}
    \begin{enumerate}
        \item \textit{Elara Vayne is the chief Star-Navigator of the starship `Wanderer'}
        \item \textit{Many young Squibs dream of joining the Sky Guard.}
        \item \textit{Jax Korden serves as the primary Rift-Warden protecting the Chronos Gate.}
    \end{enumerate}
    
    The LLM can answer the question partially using the first or third document, but to give a complete response it needs both of them.
    \end{examplebox}
    
    \item \textbf{Synergy (Multi-hop):} Answering query $Q$ requires synthesizing information from two documents $\{d_i, d_j\} \subset D$. These documents, when considered together, provide the answer, but individually offer no utility towards $R_{\mathrm{target}}$.
    \begin{examplebox}
    \textit{\textbf{Question}: What is the weather in the capital of Suvsambil?} \\[0.5em]
    \textit{\textbf{Documents}:}
    \begin{enumerate}
        \item \textit{The capital of Suvsambil is Savrak.}
        \item \textit{Weather in Tentak is cloudy.}
        \item \textit{Weather in Savrak is sunny.}
    \end{enumerate}
    
    The LLM must synthesize the capital of Suvsambil from Document 1 and then use Document 3 to answer the weather question. If either document is removed, the LLM cannot answer the question.
    \end{examplebox}
\end{itemize}

For each scenario, we built 20  query-documents pairs, using powerful LLMs such as ChatGPT, Gemini, and Deepseek, and then inspected them manually.

\paragraph{Procedure}
For each synthetic dataset scenario:
\begin{enumerate}
    \item Generate $R_{\mathrm{target}} = \mathrm{LLM}(Q, D)$.
    \item Apply all attribution methods.
    \item Analyze qualitatively the behavior of methods in inter-relation consideration for attribution computing.
\end{enumerate}

\section{Results and Analysis}
\label{sec:results}

\subsection{Experiment 1: Shapley Replication Quality}
The performance of the attribution methods was evaluated on the BIOASQ (top row) and NQ (second row) datasets. A clear and consistent performance hierarchy is observed across both datasets and all evaluation metrics.
\begin{figure}[t]
    \centering
    \includegraphics[width=0.7\textwidth]{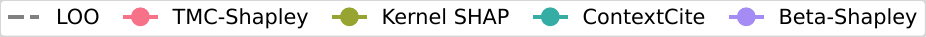}

    \begin{subfigure}[b]{0.23\textwidth}
        \includegraphics[width=\textwidth]{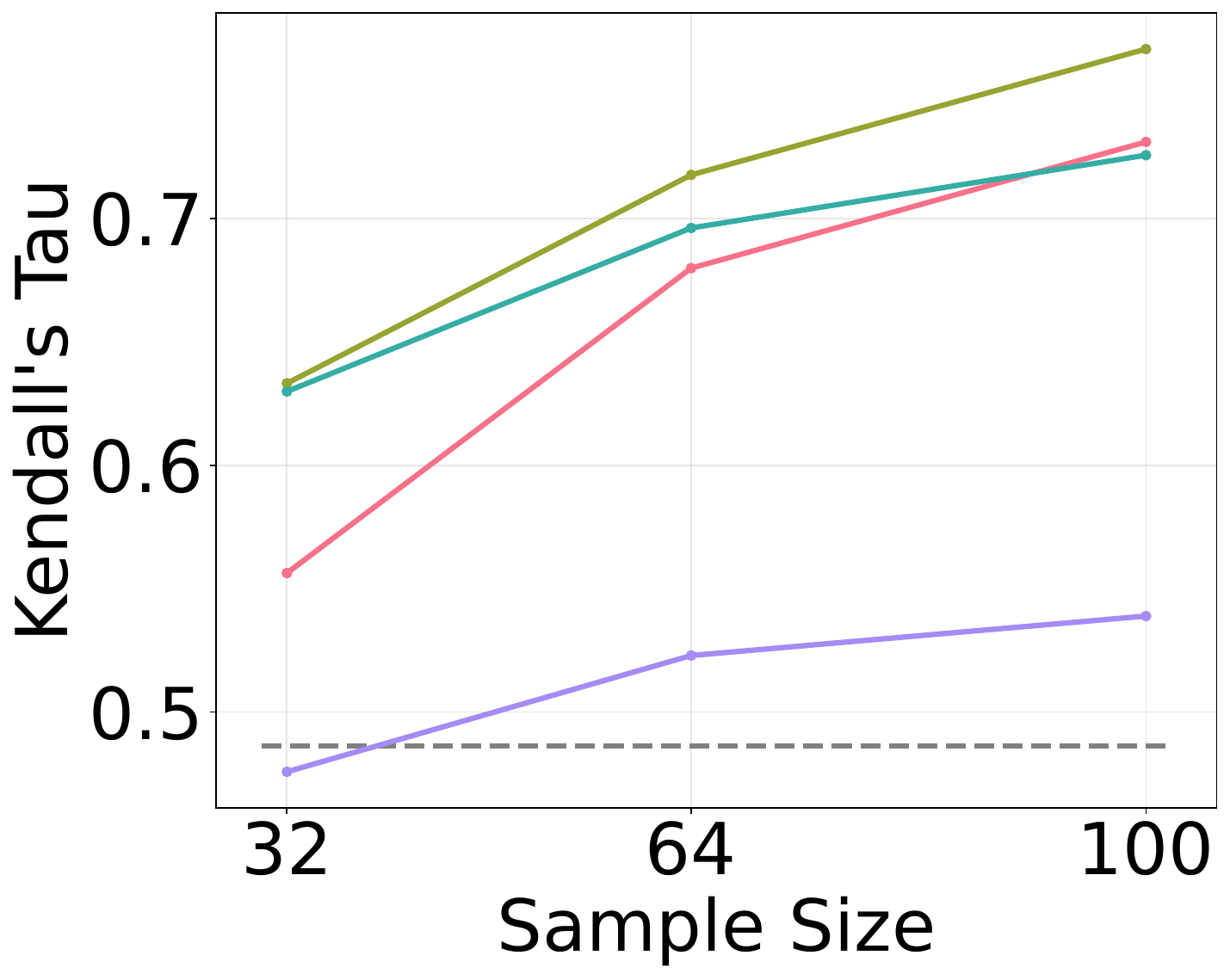}
        \label{fig:bioask_kendall}
    \end{subfigure}
    \hfill
    \begin{subfigure}[b]{0.23\textwidth}
        \includegraphics[width=\textwidth]{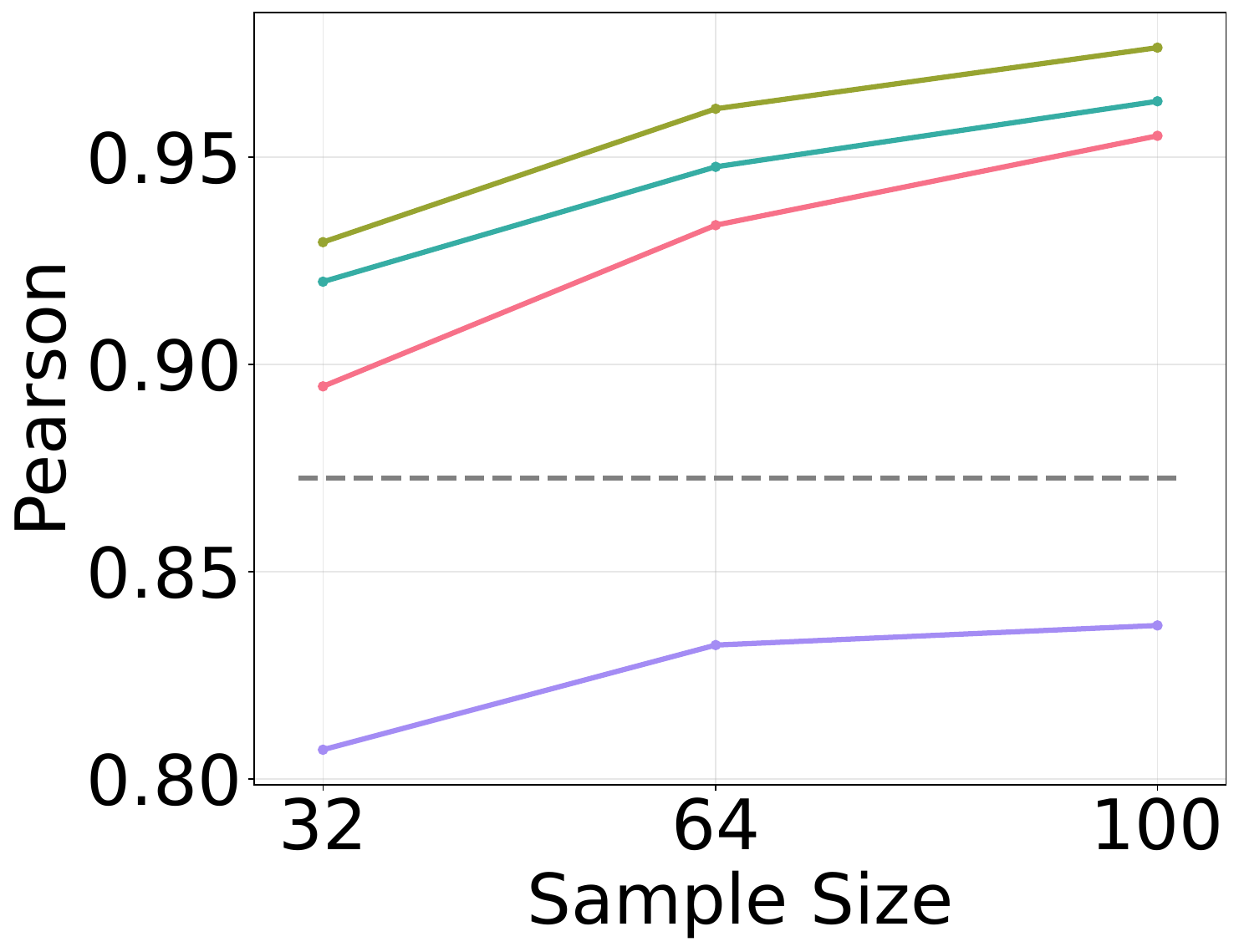}
        \label{fig:bioask_pearson}
    \end{subfigure}
    \hfill
    \begin{subfigure}[b]{0.23\textwidth}
        \includegraphics[width=\textwidth]{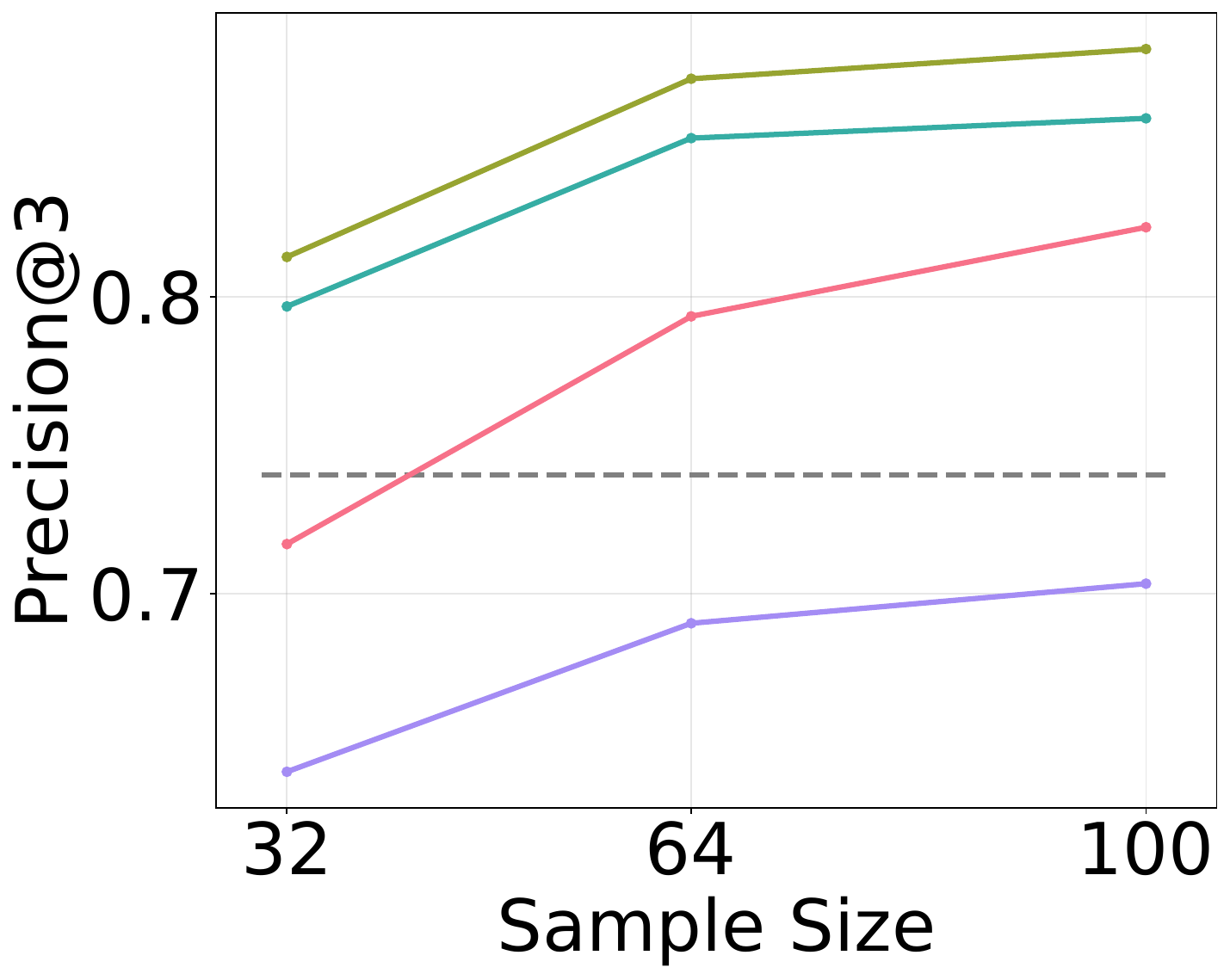}
        \label{fig:bioask_prec3}
    \end{subfigure}
    \hfill
    \begin{subfigure}[b]{0.23\textwidth}
        \includegraphics[width=\textwidth]{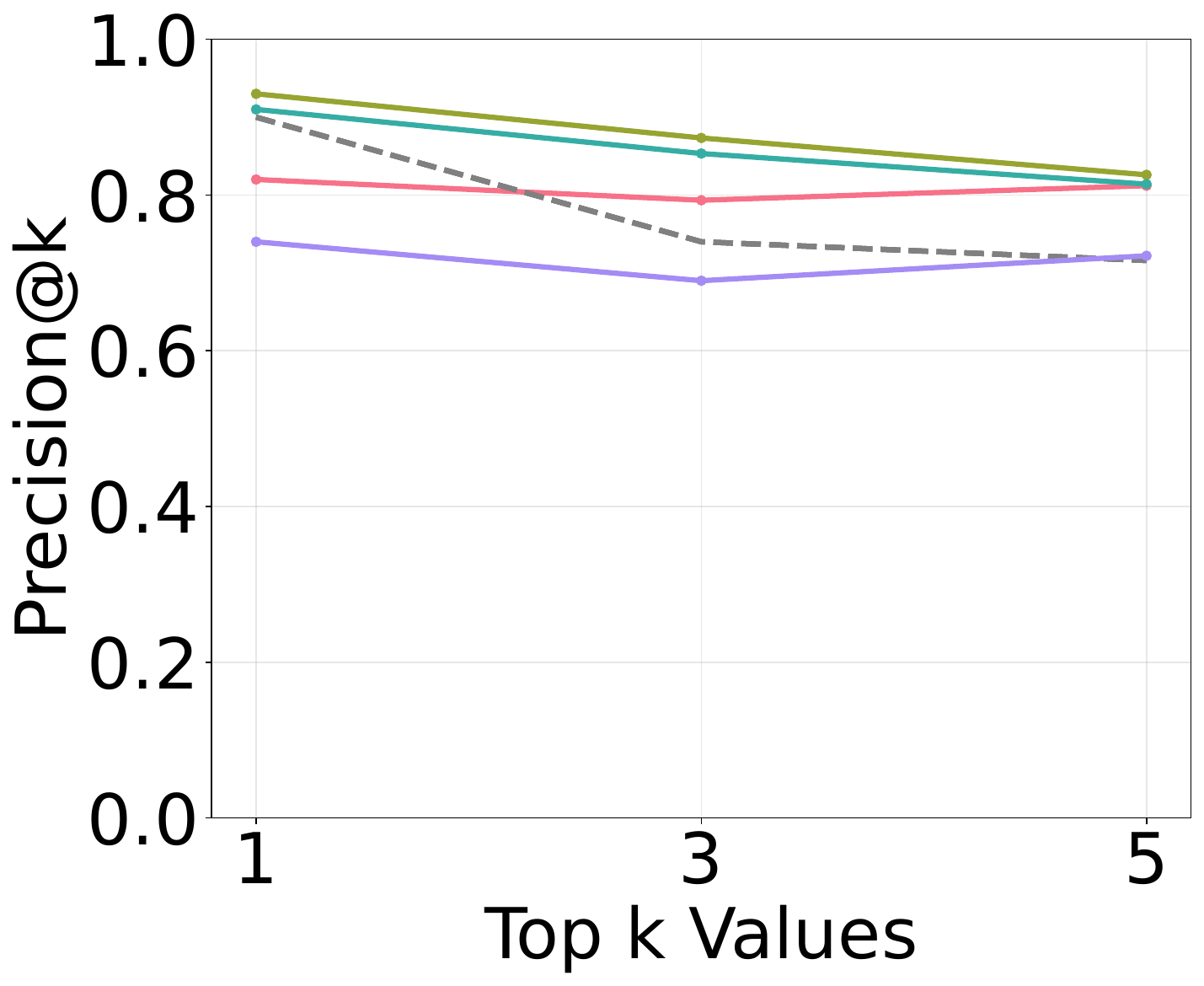}
        \label{fig:bioask_topk}
    \end{subfigure}

    \begin{subfigure}[b]{0.23\textwidth}
        \includegraphics[width=\textwidth]{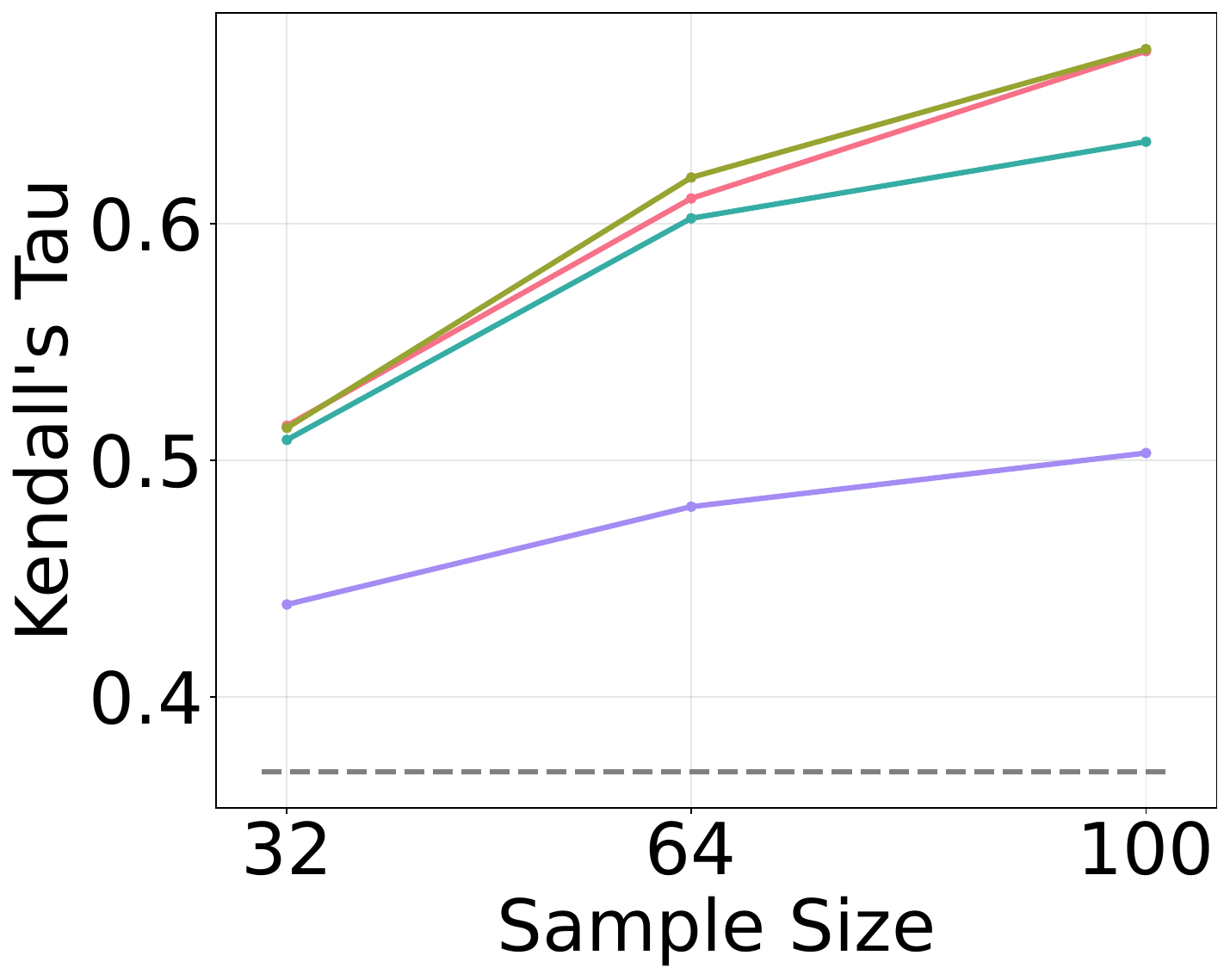}
        \label{fig:nq_kendall}
    \end{subfigure}
    \hfill
    \begin{subfigure}[b]{0.23\textwidth}
        \includegraphics[width=\textwidth]{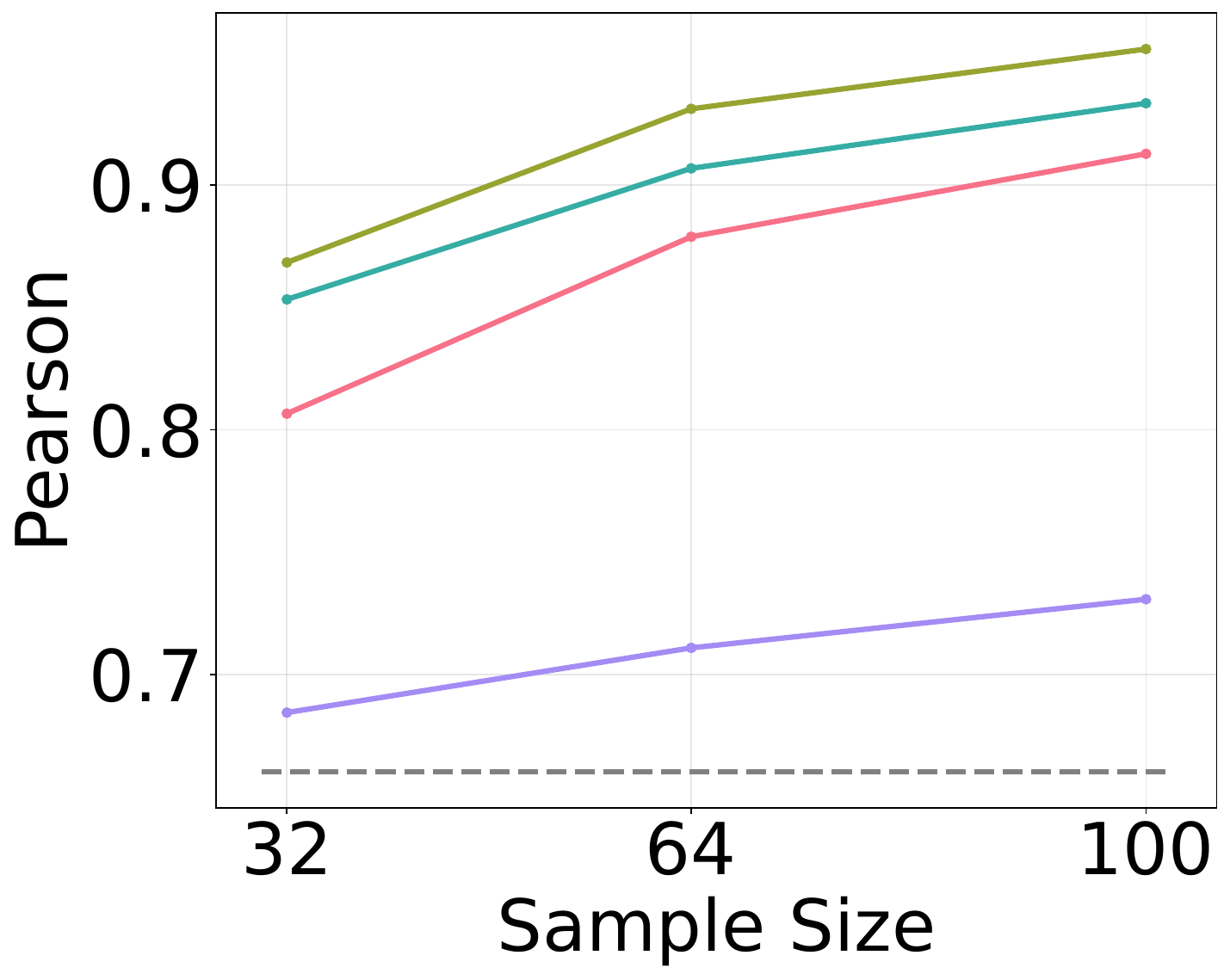}
        \label{fig:nq_pearson}
    \end{subfigure}
    \hfill
    \begin{subfigure}[b]{0.23\textwidth}
        \includegraphics[width=\textwidth]{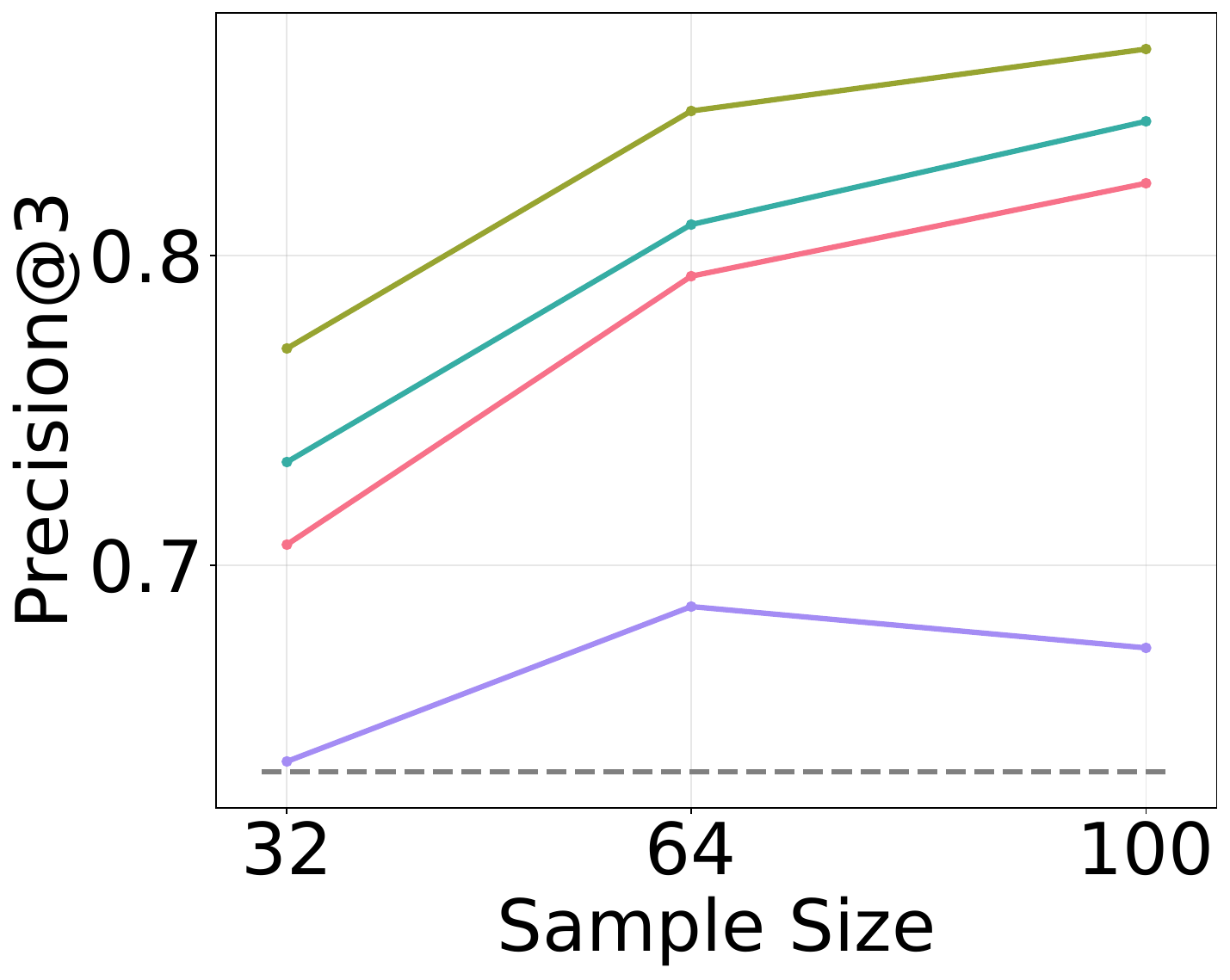}
        \label{fig:nq_prec3}
    \end{subfigure}
    \hfill
    \begin{subfigure}[b]{0.23\textwidth}
        \includegraphics[width=\textwidth]{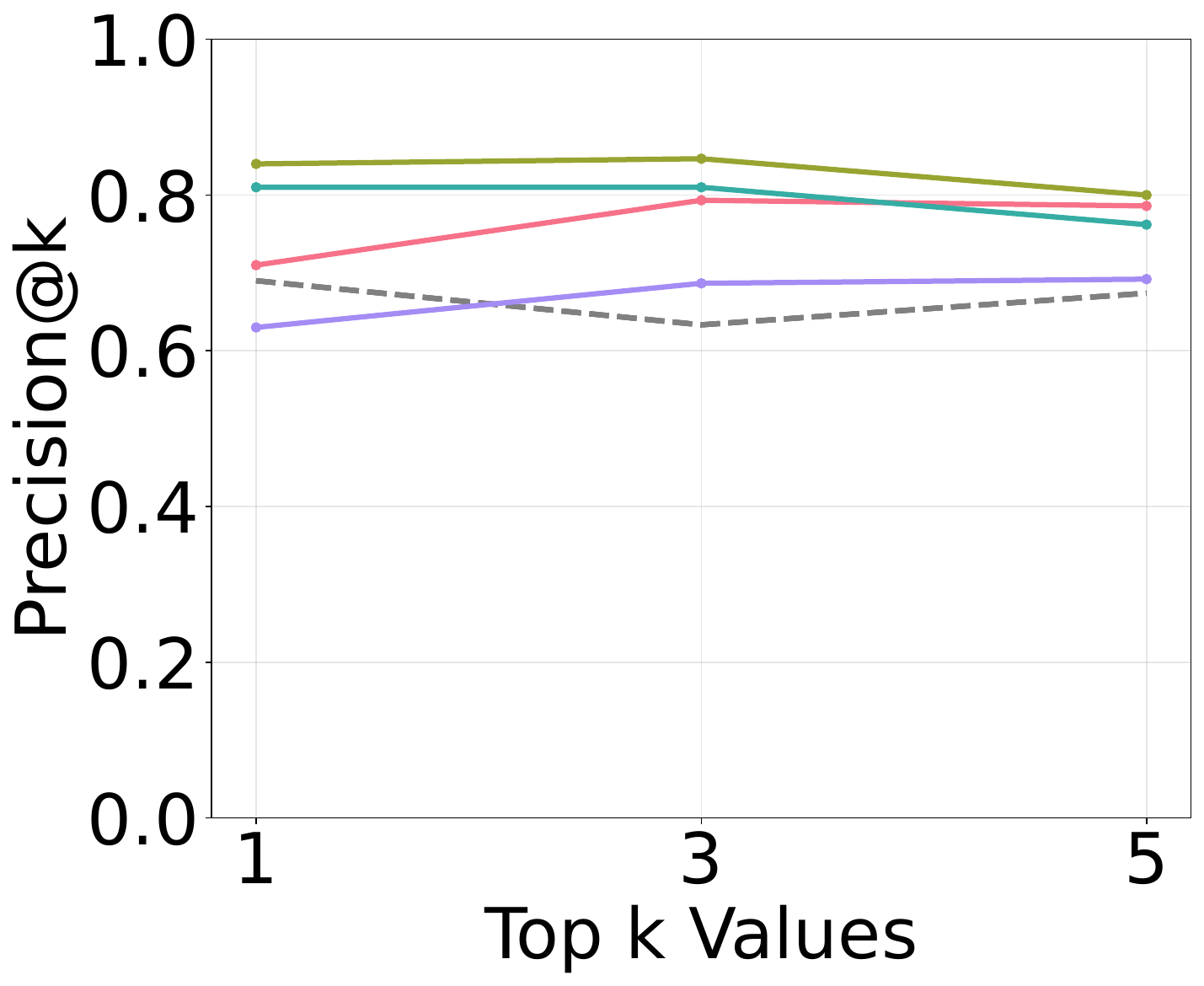}
        \label{fig:nq_topk}
    \end{subfigure}

    \caption{Experiment 1 with \textbf{Mistral}: Performance on BIOASQ and NQ.}
    \label{fig:exp1_mistral_all}
\end{figure}

Kernel SHAP consistently emerges as the top-performing attribution method, achieving the highest scores across the majority of metrics. It is closely followed by ContextCite, which also demonstrates very strong performance. Both methods, which leverage a surrogate model approach, significantly outperform the other techniques. Although TMC-Shapley and Beta Shapley aim for computationally efficient approximation, they require more ground truth utilities ($n \times T$ where $T=|D|!$ and $n$ is a constant) for high accuracy \cite{datashap}.

The impact of sample size on approximation quality is illustrated in the first three columns of each row. As the number of samples increases from 32 to 64 and finally 100, most sampling-based methods exhibit monotonic improvement.

When evaluating the identification of top-ranked documents (rightmost column), most methods are effective at identifying the single most important document (Precision@1 is generally above 0.8). When $k$ increases, the top performing methods (Kernel SHAP, ContextCite and TMC - Shapley) stay constant showing only a minimal decrease in performance for both datasets. Nevertheless, the top-performers stay ahead of the LOO and Beta-Shapley methods in all cases. 

This experiment shows that Shapley values can be effectively approximated by computationally less exhaustive methods even in RAG scenario where minimum utility computation is preferred.

\subsection{Experiment 2: Attribution Effectiveness}
\begin{table}[t]
    \centering
    \scriptsize
    \begin{tabularx}{\textwidth}{bssssssssssss}
        \toprule
         & \multicolumn{4}{l}{\textbf{Qwen 3B}} & \multicolumn{4}{l}{\textbf{Mistral 7B}} & \multicolumn{4}{l}{\textbf{Llama 8B}} \\
        $k$ & 2 & 3 & 4 & 5 & 2 & 3 & 4 & 5 & 2 & 3 & 4 & 5 \\
        \midrule
        Shapley & 0.73 & 0.70 & 0.73 & 0.74 & \textbf{0.80} & \textbf{0.77} & \textbf{0.78} & \textbf{0.78} & 0.79 & \textbf{0.78} & \textbf{0.75} & \textbf{0.78} \\ \midrule
        TMC-Shapley 32 & 0.61 & 0.58 & 0.61 & 0.66 & 0.69 & 0.67 & 0.70 & 0.70 & 0.62 & 0.61 & 0.65 & 0.67 \\
        Beta Shapley 32 &  0.53 & 0.51 & 0.58 & 0.61 & 0.61 & 0.59 & 0.61 & 0.66 & 0.55 & 0.58 & 0.61 & 0.63 \\
        Kernel SHAP 32 & 0.72 & 0.67 & 0.68 & 0.70 & 0.75 & 0.69 & 0.72 & 0.73 & 0.74 & 0.70 & 0.69 & 0.69 \\
        ContextCite 32 & 0.72 & 0.67 & 0.65 & 0.68 & 0.73 & 0.68 & 0.68 & 0.68 & 0.70 & 0.71 & 0.68 & 0.69 \\ \midrule
        TMC-Shapley 64 & 0.64 & 0.64 & 0.66 & 0.68 & 0.70 & 0.71 & 0.73 & 0.72 & 0.68 & 0.66 & 0.68 & 0.72 \\
        Beta Shapley 64 & 0.58 & 0.59 & 0.57 & 0.62 & 0.66 & 0.60 & 0.63 & 0.65 & 0.59 & 0.55 & 0.62 & 0.65 \\
        Kernel SHAP 64 & 0.74 & 0.71 & 0.73 & 0.74 & 0.77 & 0.75 & 0.76 & 0.75 & 0.79 & 0.73 & 0.74 & 0.77 \\
        ContextCite 64 & 0.74 & 0.71 & 0.72 & 0.73 & \textbf{0.80} & 0.75 & 0.73 & 0.74 & 0.78 & 0.72 & 0.71 & 0.71 \\ \midrule
        TMC-Shapley 100 & 0.68 & 0.65 & 0.70 & 0.70 & 0.76 & 0.71 & 0.74 & 0.73 & 0.72 & 0.69 & 0.71 & 0.75 \\
        Beta Shapley 100 & 0.57 & 0.56 & 0.62 & 0.65 & 0.61 & 0.61 & 0.64 & 0.66 & 0.61 & 0.56 & 0.61 & 0.65 \\
        Kernel SHAP 100 & \textbf{0.75} & 0.71 & 0.73 & \textbf{0.75} & 0.79 & 0.73 & 0.76 & 0.76 & 0.79 & 0.73 & 0.76 & 0.75 \\
        ContextCite 100 & \textbf{0.75} & \textbf{0.74} & \textbf{0.74} & 0.73 & \textbf{0.80} & \textbf{0.77} & 0.76 & 0.77 & \textbf{0.82} & 0.75 & 0.73 & 0.73 \\ \midrule
        LOO & 0.61 & 0.55 & 0.52 & 0.60 & 0.57 & 0.57 & 0.57 & 0.61 & 0.62 & 0.54 & 0.53 & 0.59 \\
        \bottomrule
    \end{tabularx}
    \caption{Exhaustive top-$k$ for the NQ dataset.}
    \label{tab:exhaustive_top_k_NQ}
\end{table}
In this experiment, we measure how effective the tested methods are at identifying documents that can cause the highest utility drop when removed. The procedure was described in Section \ref{sec:evaluation_framework} and the used metric is Precision@k ($P@k_{\mathrm{impact}}$). The results for the NQ dataset are presented in Table \ref{tab:exhaustive_top_k_NQ} and for BIOASQ could be found in the Appendix.

The Shapley method consistently achieves the highest or near-highest precision across all models and $k$ values. This aligns with expectations, as Shapley values exhaustively evaluate utilities for all $2^{|D|}$ possible document subsets. Nevertheless, KernelSHAP and ContextCite often perform comparably well because their linear models uphold the additivity axiom.

Overall, all methods demonstrate reasonable attribution precision, though some imperfections remain. We attribute this to inter-document dependencies in the datasets, which are not always fully captured by Shapley values. We examine this phenomenon in more detail in Experiment 3.


\subsection{Experiment 3: Inter-document Relationships}

\begin{figure}[H]
    \centering

    \begin{minipage}[b]{\textwidth}
        \centering
        \resizebox{\textwidth}{!}{
        \begin{tabular}{|l|p{12.5cm}|}
            \hline
            \rowcolor{lightgrey}
            \textbf{Question} & What is the traditional greeting in Blimpton? \\
            \hline
            \textbf{Positive A (Orange)} & The traditional greeting in Blimpton involves touching elbows while saying ``Flurp be with you''. \\
            \hline
            \textbf{Positive B (Blue)} & Blimptonian etiquette requires the elbow-touch greeting, accompanied by the standard phrase ``Flurp be with you''. \\
            \hline
            \textbf{Negative sample} & All Blimptonian vehicles hover at least 10cm above ground.\\
            \hline
            \textbf{Negative sample} & Blimptonian water freezes at 50°C due to added minerals. \\
            \hline
        \end{tabular}
        }
    \end{minipage}
    \begin{minipage}[b]{\textwidth}
        \centering
        \parbox{0.95\textwidth}{
        \centering
        \textbf{LLM answer (AB):} The traditional greeting in Blimpton is touching elbows while saying ``Flurp be with you''.
        }
        \includegraphics[width=0.9\textwidth]{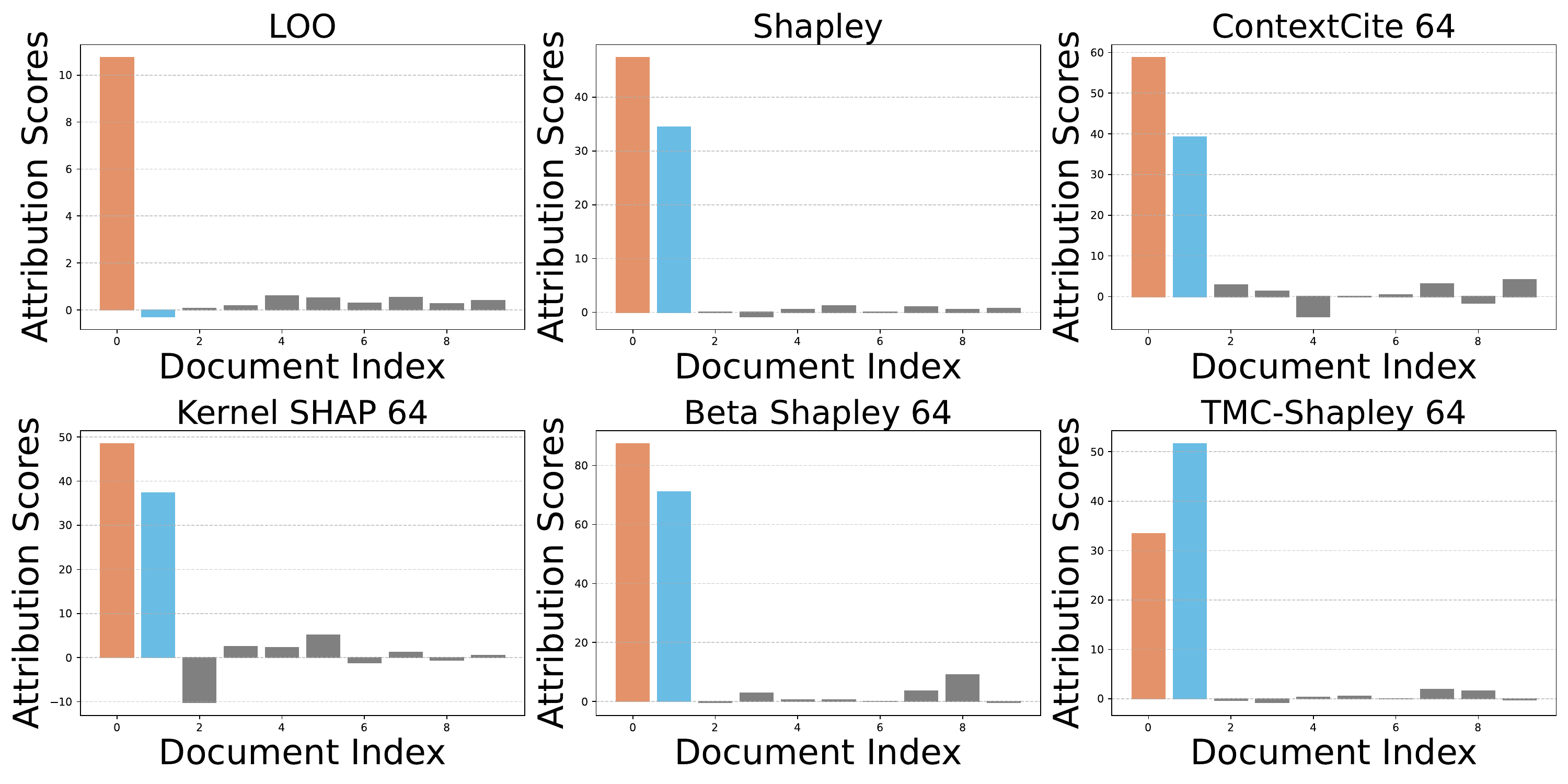}
        \hfill
        \centering
        \parbox{0.95\textwidth}{
        \textbf{LLM answer (BA):} The traditional greeting in Blimpton is the elbow-touch greeting, accompanied by the standard phrase ``Flurp be with you''.
        }
        \includegraphics[width=0.9\textwidth]{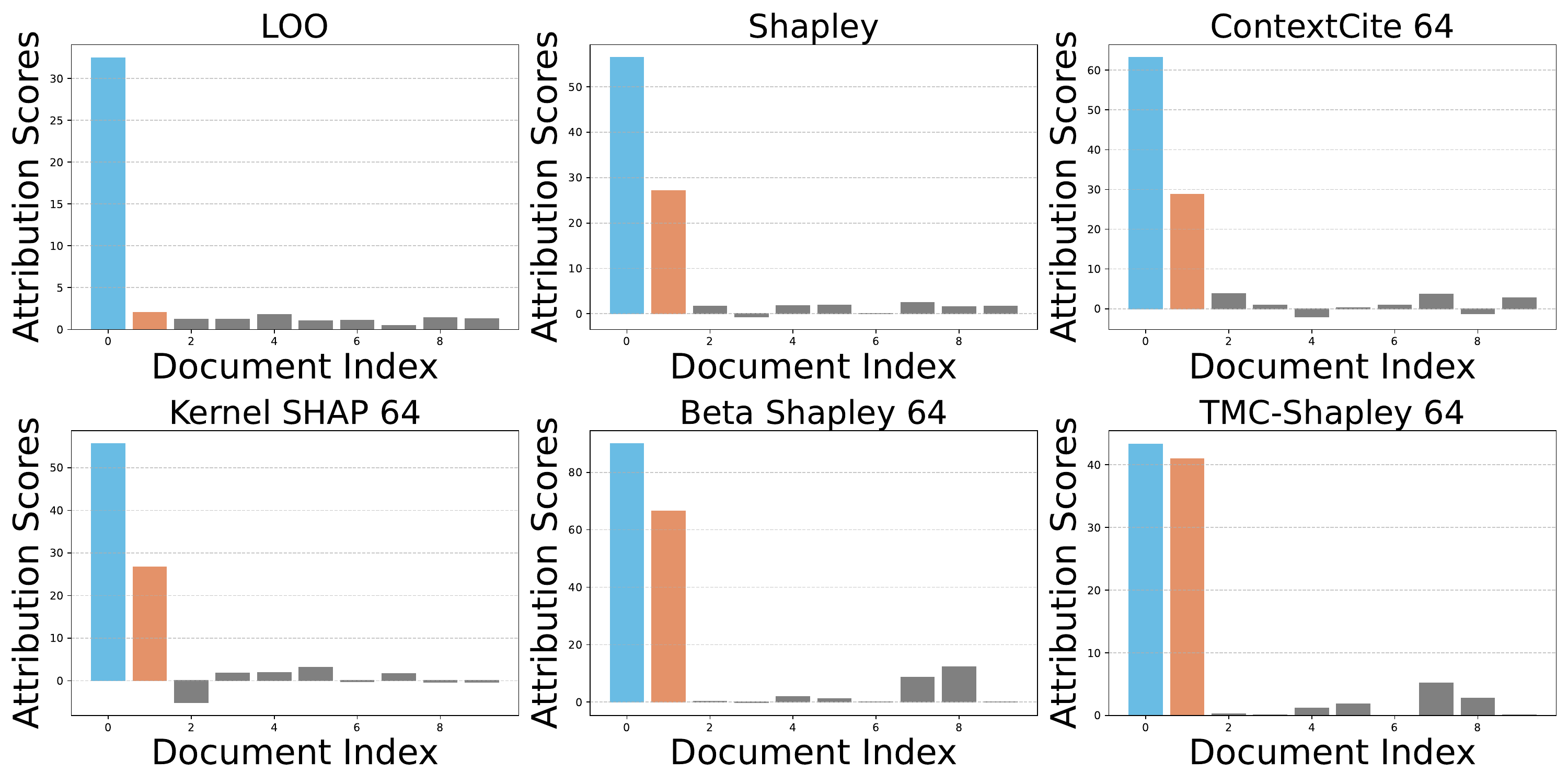}
    \end{minipage}

    \caption{(Top) Sample question belonging to the redundancy scenario (Bottom) Attribution scores visualizations for the two orderings (AB and BA).}
    \label{fig:duplicate_anecdote}
\end{figure}
In this experiment, we thoroughly examine the behavior of the attribution methods in scenarios involving complex inter-relations between documents.

\begin{figure}[H]
    \centering

    \begin{minipage}[b]{\textwidth}
        \centering
        \resizebox{\textwidth}{!}{
        \begin{tabular}{|l|p{12.5cm}|}
            \hline
            \rowcolor{lightgrey}
            \textbf{Question} & What two functions does the Mystic ``Dream-Weaver's Loom'' perform? \\
            \hline
            \textbf{Positive A (Orange)} & The Mystic ``Dream-Weaver's Loom'' can capture and solidify ``Sleep-Visions'' into physical dream-silk tapestries. \\
            \hline
            \textbf{Positive B (Blue)} & It also has the ability to ``Memory-Stitch'', embedding specific recollections directly into the fabric. \\
            \hline
            \textbf{Negative sample} & Mystic looms are powered by lunar energy.\\
            \hline
            \textbf{Negative sample} & Memory-Stitching requires deep meditative states \\
            \hline
        \end{tabular}
        }
    \end{minipage}
    \begin{minipage}[b]{\textwidth}
        \centering
        \parbox{0.95\textwidth}{
        \centering
        \textbf{LLM answer (AB):}  Mystic ``Dream-Weaver's Loom'' performs two functions.
        }
        \includegraphics[width=0.9\textwidth]{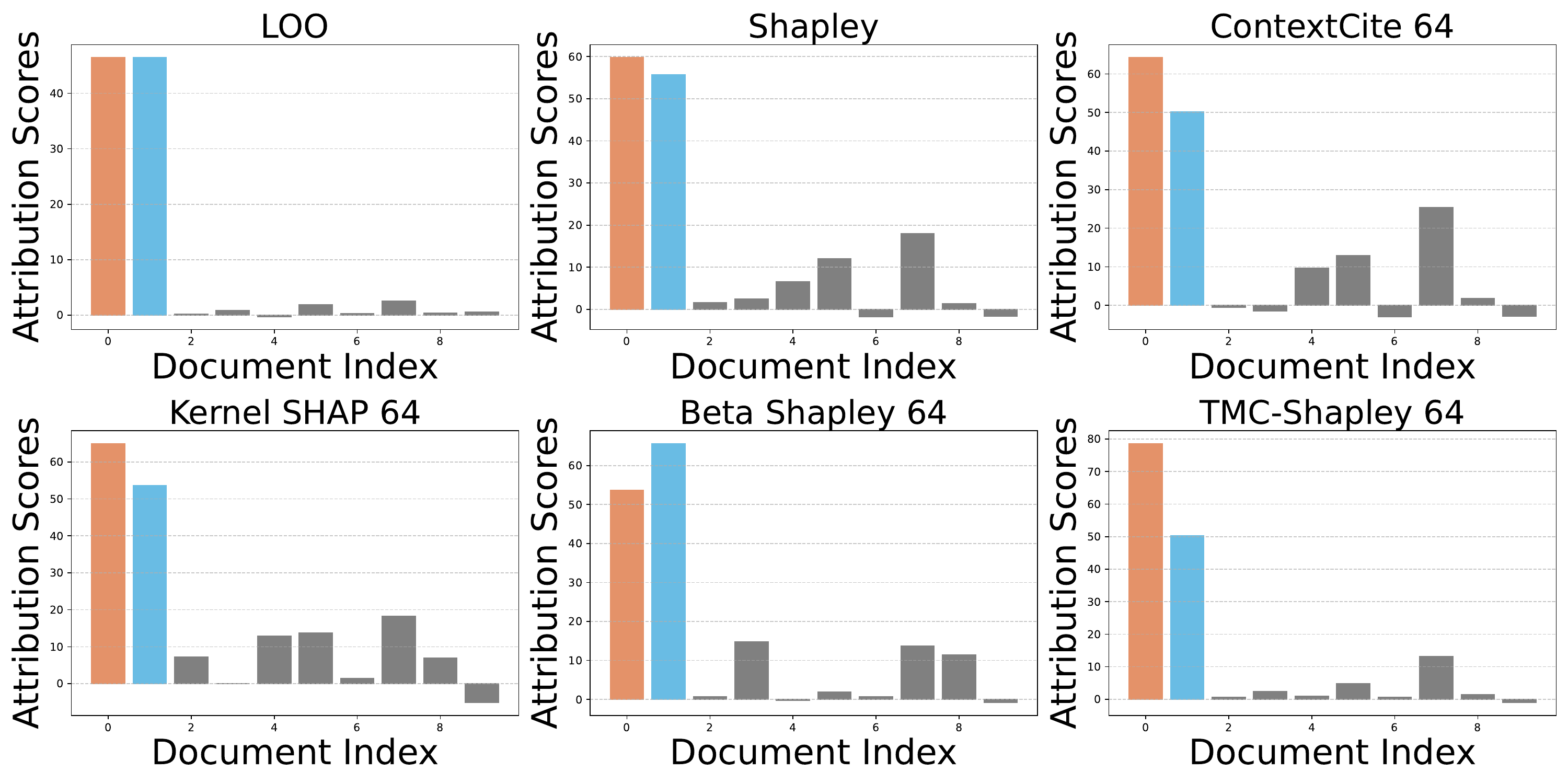}
        \hfill
        \centering
        \parbox{0.95\textwidth}{
        \textbf{LLM answer (BA):} Mystic ``Dream-Weaver's Loom'' performs two functions.
        }
        \includegraphics[width=0.9\textwidth]{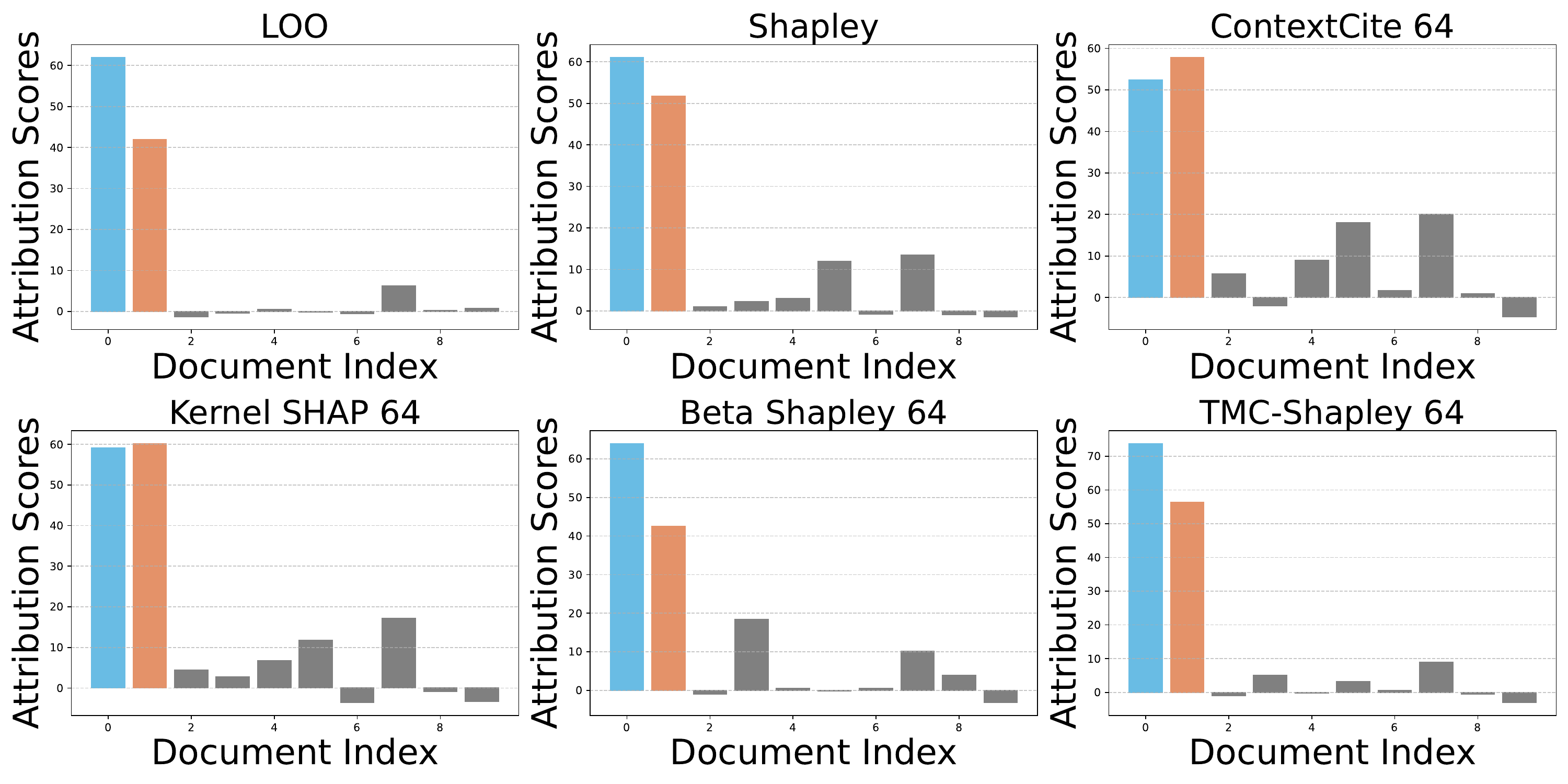}
    \end{minipage}

    \caption{(Top) Sample question belonging to the complementary scenario (Bottom) Attribution scores visualizations for the two orderings (AB and BA).}
    \label{fig:complementary_anecdote}
\end{figure}

\begin{figure}[H]
    \centering

    \begin{minipage}[b]{\textwidth}
        \centering
        \resizebox{\textwidth}{!}{
        \begin{tabular}{|l|p{12.5cm}|}
            \hline
            \rowcolor{lightgrey}
            \textbf{Question} & What is the salary of the most popular actor on the planet Aethelon? \\
            \hline
            \textbf{Positive A (Orange)} & Lyra Vael is widely considered the most popular actor currently working in Aethelon's film and stage sectors. \\
            \hline
            \textbf{Positive B (Blue)} & Lyra Vael commands a salary of approximately 50 million Credits per major project, making her one of the highest earners. \\
            \hline
            \textbf{Negative sample} & Aethelon's entertainment industry is renowned for its emotionally resonant dramas and intricate historical epics. \\
            \hline
            \textbf{Negative sample} & Actors on Aethelon undergo rigorous psychological training to fully embody complex characters. \\
            \hline
        \end{tabular}
        }
    \end{minipage}
    \begin{minipage}[b]{\textwidth}
        \centering
        \parbox{0.95\textwidth}{
        \centering
        \textbf{LLM answer (AB):}  The most popular actor on the planet Aethelon, Lyra Vael, commands a salary of approximately 50 million Credits per major project.
        }
        \includegraphics[width=0.9\textwidth]{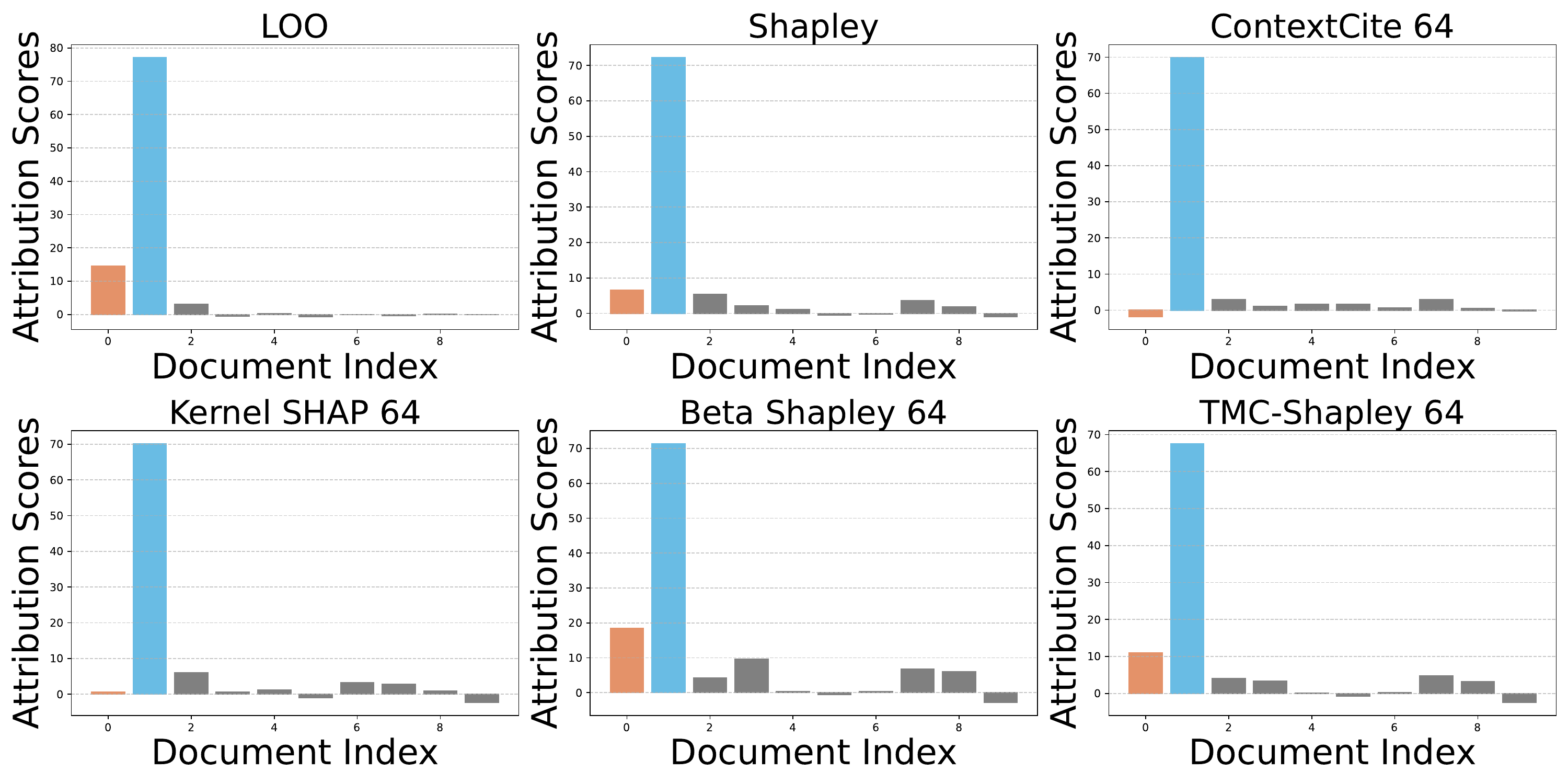}
        \hfill
        \centering
        \parbox{0.95\textwidth}{
        \textbf{LLM answer (BA):} The most popular actor on the planet Aethelon, Lyra Vael, commands a salary of approximately 50 million Credits per major project.
        }
        \includegraphics[width=0.9\textwidth]{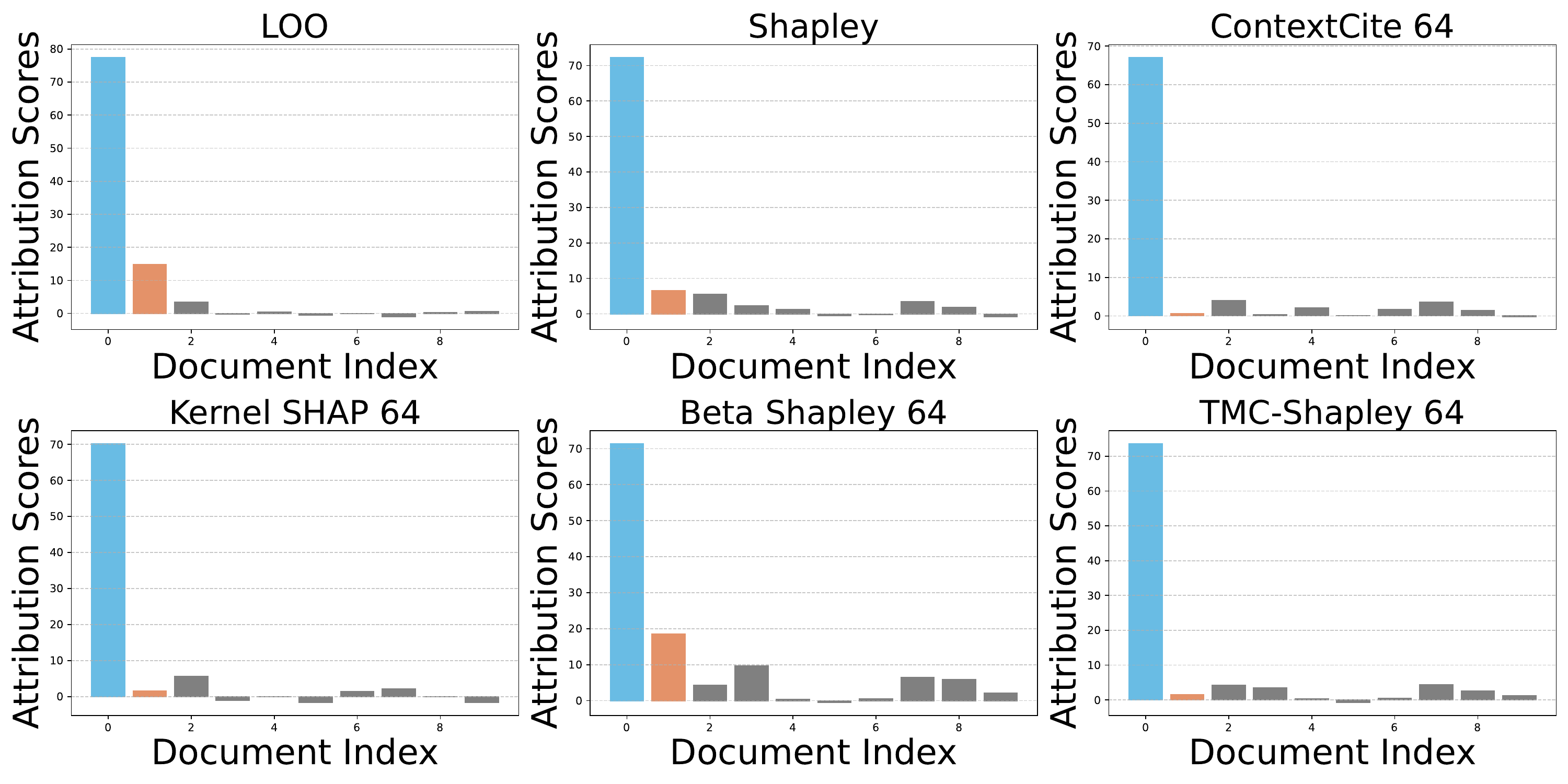}
    \end{minipage}

    \caption{(Top) Sample question belonging to the synergy scenario (Bottom) Attribution scores visualizations for the two ordering (AB and BA).}
    \label{fig:synergy_anecdote}
\end{figure}

We queried \textbf{Mistral-7B-Instruct} using questions from each scenario concatenated with relevant contextual information. The context consists of ten documents: two positive documents (A and B) placed in the first two positions, followed by eight negative documents. Each question was submitted twice to LLM, with the order of the positive documents swapped. Figures \ref{fig:duplicate_anecdote}, \ref{fig:complementary_anecdote}, and \ref{fig:synergy_anecdote} show the question, the positive documents, two negative documents, the model responses, and the calculated attribution scores for both query variants.

In the redundancy scenario, displayed in Figure \ref{fig:duplicate_anecdote}, where two near-duplicate documents independently suffice to answer the question, we observed a notable position bias: the duplicate appearing first consistently receives a higher attribution score. This effect persists even when the order of the duplicates is swapped. In Figure \ref{fig:duplicate_anecdote}, documents A and B are duplicates; A receives a higher score when it precedes B, and vice versa when their order is reversed.

This position bias is also evident in the complementarity scenario shown in Figure \ref{fig:complementary_anecdote}. All methods perform well here, correctly identifying both complementary documents, matching the exhaustive search baseline.
It is important to note that, when relying solely on the utility values of the documents, it becomes impossible to distinguish between scenarios involving complementary documents and those involving duplicate documents. In both cases, one positive document exhibits a high utility while the other shows a marginally lower utility.

In the synergy scenario presented in Figure \ref{fig:synergy_anecdote}, where the answer requires synthesis of information, first by inferring from one document and then answering the question with another, an ideal method should assign high scores to both documents. However, as shown in the figure, all methods disproportionately favor the document containing the direct answer, while underestimating the contribution of the document required for synthesis. To ensure this effect is not due to position bias, we swapped the order of the documents and found that the same direct-answer document continued to receive the higher score.

\begin{figure}[t]
    \centering
    \includegraphics[width=0.32\linewidth]{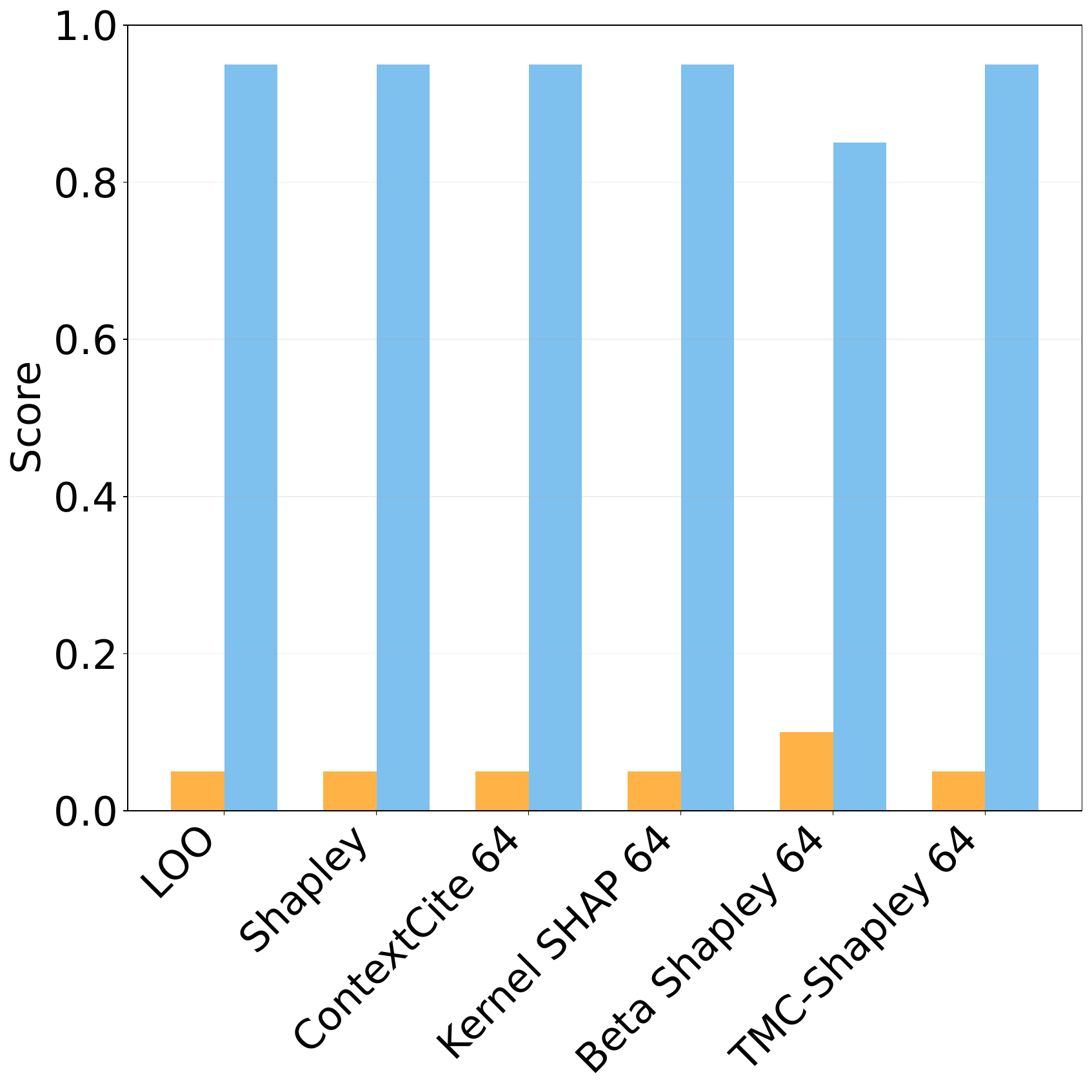}
    \includegraphics[width=0.32\linewidth]{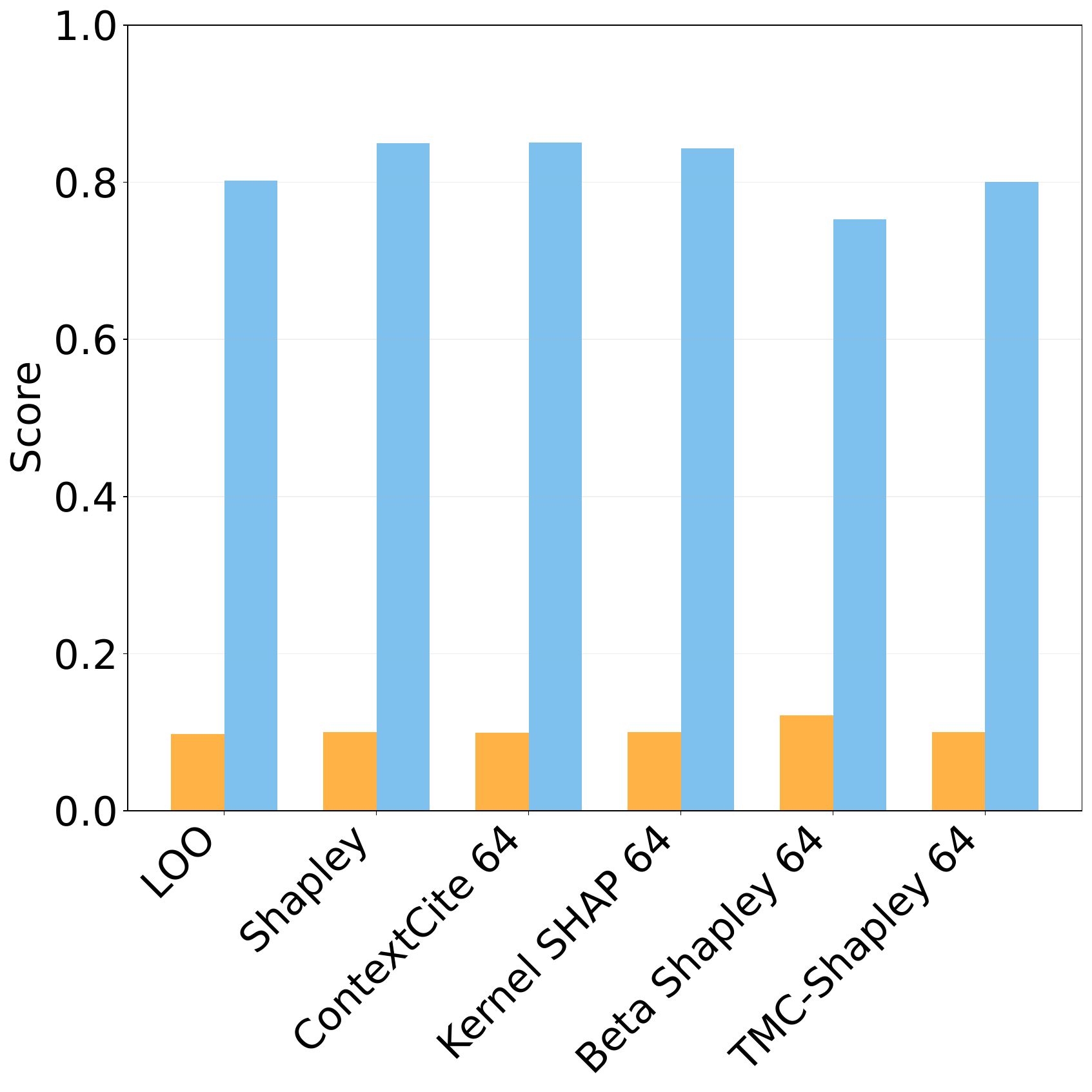}
    \includegraphics[width=0.32\linewidth]{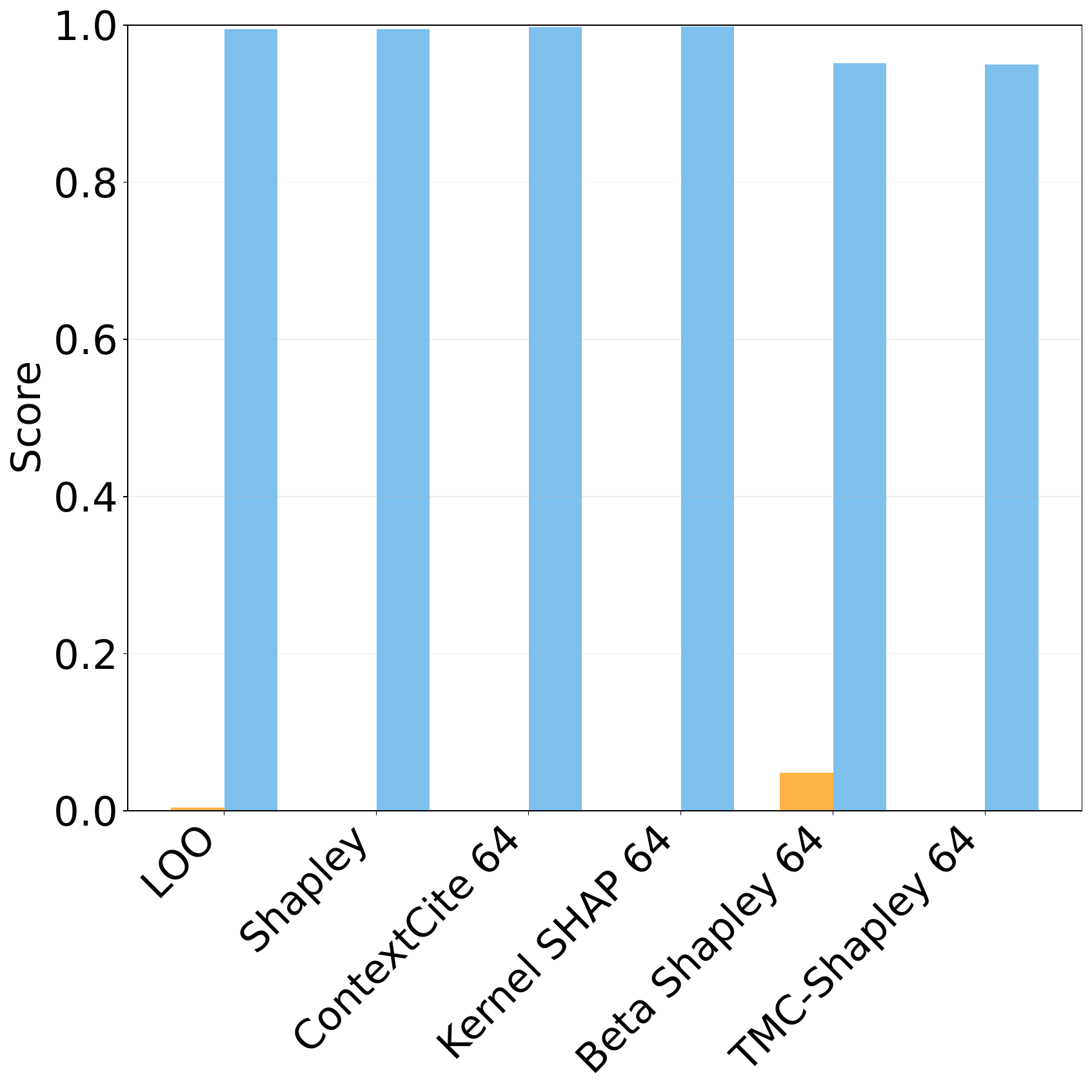}
    \caption{Average normalized attribution scores for synergetic documents A and B, tested on \textbf{Qwen-3B-Instruct}, \textbf{Mistral-7B-Instruct}, and \textbf{LLaMA-3.2-8B-Instruct}, respectively.}
    \label{fig:synergy_average}
\end{figure}

To validate that this phenomenon occurs systematically for all query-documents pairs, we normalize the attribution scores for ``A'' and ``B'' per query and then average them. As shown in Figure \ref{fig:synergy_average}, ``B'' systematically gets a much higher value than ``A''. This means the methods are undermining the contribution of the first document, which plays a crucial role in synthesizing the answer.

\section{Conclusion}
\label{sec:conclusion}
In conclusion, applying utility-based attribution methods to RAG with a tailored utility function shows promise. However, more advanced methods that account for deeper inter-document relationships are needed to improve attribution quality. As demonstrated in the final experiment, attribution scores alone may be insufficient; interpreting the interactions among highly attributed documents is also essential for a more complete understanding.

\bibliographystyle{splncs04}
\bibliography{bibliography}

\begin{thebibliography}{10}
\providecommand{\url}[1]{\texttt{#1}}
\providecommand{\urlprefix}{URL }
\providecommand{\doi}[1]{https://doi.org/#1}

\bibitem{basu2021influence}
Basu, S., Pope, P., Feizi, S.: Influence functions in deep learning are fragile. In: International Conference on Learning Representations (2021), \url{https://openreview.net/forum?id=xHKVVHGDOEk}

\bibitem{contextcite}
Cohen-Wang, B., Shah, H., Georgiev, K., M\k{a}dry, A.: Contextcite: Attributing model generation to context. In: Globerson, A., Mackey, L., Belgrave, D., Fan, A., Paquet, U., Tomczak, J., Zhang, C. (eds.) Advances in Neural Information Processing Systems. vol.~37, pp. 95764--95807. Curran Associates, Inc. (2024), \url{https://proceedings.neurips.cc/paper_files/paper/2024/file/adbea136219b64db96a9941e4249a857-Paper-Conference.pdf}

\bibitem{datashap}
Ghorbani, A., Zou, J.: Data shapley: Equitable valuation of data for machine learning. In: Chaudhuri, K., Salakhutdinov, R. (eds.) Proceedings of the 36th International Conference on Machine Learning. Proceedings of Machine Learning Research, vol.~97, pp. 2242--2251. PMLR (09--15 Jun 2019), \url{https://proceedings.mlr.press/v97/ghorbani19c.html}

\bibitem{seat}
Hu, L., Liu, Y., Liu, N., Huai, M., Sun, L., Wang, D.: Seat: stable and explainable attention. In: Proceedings of the Thirty-Seventh AAAI Conference on Artificial Intelligence and Thirty-Fifth Conference on Innovative Applications of Artificial Intelligence and Thirteenth Symposium on Educational Advances in Artificial Intelligence. AAAI'23/IAAI'23/EAAI'23, AAAI Press (2023). \doi{10.1609/aaai.v37i11.26517}, \url{https://doi.org/10.1609/aaai.v37i11.26517}

\bibitem{wra}
Huang, Y., Jia, A., Zhang, X., Zhang, J.: Generic attention-model explainability by weighted relevance accumulation. In: Proceedings of the 5th ACM International Conference on Multimedia in Asia. MMAsia '23, Association for Computing Machinery, New York, NY, USA (2024). \doi{10.1145/3595916.3626437}, \url{https://doi.org/10.1145/3595916.3626437}

\bibitem{datamodels}
Ilyas, A., Engstrom, L., Santurkar, S., Tsipras, D., Madry, A.: Data models: Understanding data through models. In: Proceedings of the 39th International Conference on Machine Learning (ICML). pp. 9525--9589. PMLR (2022), \url{https://proceedings.mlr.press/v162/ilyas22a.html}

\bibitem{jain-wallace-2019-attention}
Jain, S., Wallace, B.C.: {A}ttention is not {E}xplanation. In: Burstein, J., Doran, C., Solorio, T. (eds.) Proceedings of the 2019 Conference of the North {A}merican Chapter of the Association for Computational Linguistics: Human Language Technologies, Volume 1 (Long and Short Papers). pp. 3543--3556. Association for Computational Linguistics, Minneapolis, Minnesota (Jun 2019). \doi{10.18653/v1/N19-1357}, \url{https://aclanthology.org/N19-1357/}

\bibitem{if}
Koh, P.W., Liang, P.: Understanding black-box predictions via influence functions. In: Precup, D., Teh, Y.W. (eds.) Proceedings of the 34th International Conference on Machine Learning. Proceedings of Machine Learning Research, vol.~70, pp. 1885--1894. PMLR (06--11 Aug 2017), \url{https://proceedings.mlr.press/v70/koh17a.html}

\bibitem{nq}
Kwiatkowski, T., Palomaki, J., Redfield, O., Collins, M., Parikh, A., Alberti, C., Epstein, D., Polosukhin, I., Devlin, J., Lee, K., et~al.: Natural questions: a benchmark for question answering research. Transactions of the Association for Computational Linguistics  \textbf{7},  453--466 (2019)

\bibitem{betashap}
Kwon, Y., Zou, J.: Beta shapley: a unified and noise-reduced data valuation framework for machine learning. In: Camps-Valls, G., Ruiz, F.J.R., Valera, I. (eds.) AISTATS. Proceedings of Machine Learning Research, vol.~151, pp. 8780--8802. PMLR (2022)

\bibitem{kernel-shap}
Lundberg, S.M., Lee, S.I.: A unified approach to interpreting model predictions. In: Guyon, I., Luxburg, U.V., Bengio, S., Wallach, H., Fergus, R., Vishwanathan, S., Garnett, R. (eds.) Advances in Neural Information Processing Systems. vol.~30. Curran Associates, Inc. (2017), \url{https://proceedings.neurips.cc/paper_files/paper/2017/file/8a20a8621978632d76c43dfd28b67767-Paper.pdf}

\bibitem{sarti-etal-2023-inseq}
Sarti, G., Feldhus, N., Sickert, L., van~der Wal, O.: Inseq: An interpretability toolkit for sequence generation models. In: Bollegala, D., Huang, R., Ritter, A. (eds.) Proceedings of the 61st Annual Meeting of the Association for Computational Linguistics (Volume 3: System Demonstrations). pp. 421--435. Association for Computational Linguistics, Toronto, Canada (Jul 2023). \doi{10.18653/v1/2023.acl-demo.40}, \url{https://aclanthology.org/2023.acl-demo.40/}

\bibitem{shapley1953}
Shapley, L.S., et~al.: A value for n-person games  (1953)

\bibitem{Simonyan2013Deep}
Simonyan, K., Vedaldi, A., Zisserman, A.: Deep inside convolutional networks: Visualising image classification models and saliency maps. arXiv preprint arXiv:1312.6034  (2013)

\bibitem{Smilkov2017SmoothGrad}
Smilkov, D., Thorat, N., Kim, B., Vi{\'e}gas, F., Wattenberg, M.: Smoothgrad: removing noise by adding noise. arXiv preprint arXiv:1706.03825  (2017)

\bibitem{Sundararajan2017Axiomatic}
Sundararajan, M., Taly, A., Yan, Q.: Axiomatic attribution for deep networks. In: Precup, D., Teh, Y.W. (eds.) Proceedings of the 34th International Conference on Machine Learning. Proceedings of Machine Learning Research, vol.~70, pp. 3319--3328. PMLR (06--11 Aug 2017), \url{https://proceedings.mlr.press/v70/sundararajan17a.html}

\bibitem{bioasq}
Tsatsaronis, G., Balikas, G., Malakasiotis, P., Partalas, I., Zschunke, M., Alvers, M.R., Weissenborn, D., Krithara, A., Petridis, S., Polychronopoulos, D., et~al.: An overview of the bioasq large-scale biomedical semantic indexing and question answering competition. BMC bioinformatics  \textbf{16},  1--28 (2015)

\bibitem{Vaswani2017Attention}
Vaswani, A., Shazeer, N., Parmar, N., Uszkoreit, J., Jones, L., Gomez, A.N., Kaiser, L.u., Polosukhin, I.: Attention is all you need. In: Guyon, I., Luxburg, U.V., Bengio, S., Wallach, H., Fergus, R., Vishwanathan, S., Garnett, R. (eds.) Advances in Neural Information Processing Systems. vol.~30. Curran Associates, Inc. (2017), \url{https://proceedings.neurips.cc/paper_files/paper/2017/file/3f5ee243547dee91fbd053c1c4a845aa-Paper.pdf}

\bibitem{wiegreffe-pinter-2019-attention}
Wiegreffe, S., Pinter, Y.: Attention is not not explanation. In: Inui, K., Jiang, J., Ng, V., Wan, X. (eds.) Proceedings of the 2019 Conference on Empirical Methods in Natural Language Processing and the 9th International Joint Conference on Natural Language Processing (EMNLP-IJCNLP). pp. 11--20. Association for Computational Linguistics, Hong Kong, China (Nov 2019). \doi{10.18653/v1/D19-1002}, \url{https://aclanthology.org/D19-1002/}

\bibitem{contrastllm}
Yin, K., Neubig, G.: Interpreting language models with contrastive explanations

\end{thebibliography}

\newpage
\appendix
\section{Model Evaluation}
We employ two evaluation metrics to evaluate the quality of responses by LLMs:  
\begin{enumerate}
    \item \textbf{Cosine similarity} between the embeddings of the generated and ground-truth answers. For the BioASQ dataset, we use \texttt{MedEmbed-large-v0.1}; for all other datasets, we use \texttt{all-MiniLM-L6-v2}, both from the Sentence Transformers library.
    \item \textbf{LLM-as-a-Judge} - Given the question, generated and ground-truth answers, we use a fine-tuned LLM to decide whether the generated response entails the ground truth. The judge returns a binary output, True or False.
\end{enumerate}


Evaluation of the models for all datasets is demonstrated in the table. The results suggest that cosine similarity can be a misleading metric for factual accuracy.  This is highlighted by the conflicting trends between the two evaluation methods.

For the BIOASQ dataset, cosine scores (0.87-0.88) are high, while for the NQ dataset, they are very much lower (0.38-0.42). A likely explanation for this discrepancy is the nature of the ground-truth answers in each dataset. NQ's ground truths are often very short (one or a few words), while BIOASQ's are typically complete sentences. The brevity of NQ answers provides very little semantic information for a cosine similarity metric, resulting in low scores. In contrast, the longer sentences in BIOASQ offer a much richer context, making it easier to find high semantic similarity and achieve high scores.

Regarding the synthetic datasets, the models demonstrate exceptionally strong performance. The LLM-judge scores are particularly revealing, showing near-perfect capabilities: models scored a perfect 1.0 for Redundancy, 0.95-1.0 for Synergy, and 0.85-1.0 for Complementarity. The cosine similarity scores, while lower, also point to a strong ability to handle these tasks (ranging from 0.75 to 0.91). This indicates a robust capacity to synthesize information from multiple sources, whether it is redundant, complementary, or requires deeper reasoning.

The LLM-as-a-judge used \textit{Gemini-2.0-Flash} as a base with the following system prompt: 

\begin{tcblisting}{listing only, title=System Prompt}

You are an expert judge evaluating whether two sentences are equivalent in meaning, 
both are answers to the same query. One is a generated answer and the other is the ground truth.

Evaluation Criteria:
1. Focus on semantic equivalence, not exact wording
2. Minor grammatical differences don't affect equivalence
3. The generated answer must capture all key information from the ground truth
4. Additional relevant information in the generated answer is acceptable

Output Format (strictly follow this JSON format):
{"evaluation": "yes"/"no", "explanation": "short explanation about the provided evaluation"}

Examples:
Query: "What is photosynthesis?"
Ground Truth: "Photosynthesis is how plants make food using sunlight."
Generated Answer: "The process by which plants convert sunlight into food is called photosynthesis."
Output: {"evaluation": "yes", "explanation": "Both sentences describe the same process with equivalent meaning, though worded differently."}

Query: "Who wrote Romeo and Juliet?"
Ground Truth: "William Shakespeare wrote Romeo and Juliet."
Generated Answer: "Romeo and Juliet was a play by Shakespeare."
Output: {"evaluation": "yes", "explanation": "Both identify Shakespeare as the author, despite slight wording differences."}

Query: "What causes seasons?"
Ground Truth: "Earth's axial tilt causes seasons."
Generated Answer: "The changing distance from the sun causes seasons."
Output: {"evaluation": "no", "explanation": "The answers provide different scientific explanations for seasons."}
\end{tcblisting}

\begin{table}[t]
    \centering
    \begin{tabularx}{\textwidth}{g|s|s|s|s|s|s}
        \toprule
        & \multicolumn{2}{|c|}{\textbf{Qwen 3B}} & \multicolumn{2}{c|}{\textbf{Mistral 7B}} & \multicolumn{2}{c}{\textbf{Llama 8B}} \\
        \midrule
        & Cosine & LLM-judge & Cosine & LLM-judge & Cosine & LLM-judge \\ \midrule
        \textbf{BioAsq} & 0.87 & 0.73 & 0.88 & 0.84 & 0.87 & 0.75 \\
        \textbf{NQ} & 0.38 & 0.64 & 0.38 & 0.72 & 0.42 & 0.72 \\ \midrule
        \textbf{Redundancy} & 0.76 & 1.0 & 0.75 & 1.0 & 0.81 & 1.0 \\
        \textbf{Complementarity} & 0.90 & 0.85 & 0.90 & 1.0 & 0.91 & 0.90 \\
        \textbf{Synergy} & 0.84 & 1.0 & 0.83 & 1.0 & 0.82 & 0.95 \\
        \bottomrule
    \end{tabularx}
    \caption{Evaluation of LLMs}
    \label{tab:llm_eval}
\end{table}

\section{Additional Results}
\subsection{Experiment 1}
The relative ranking and performance trends of the methods are remarkably consistent across both datasets, though absolute scores are slightly higher on BIOASQ compared to NQ. In conclusion, the results provide strong evidence that Kernel SHAP is the most accurate and sample-efficient method for approximating Shapley values in this setting, with ContextCite being a highly competitive alternative. Both methods represent a substantial improvement in accuracy and reliability over TMC-Shapley, Beta-Shapley, and LOO.

Across both datasets: BIOASQ (first row) and  NQ (second row) and across all models, Kernel SHAP consistently stands out as the best method, achieving the highest scores across all metrics. It is closely followed by ContextCite which also demonstrates strong performance. Both methods use the surrogate model approach (use the weights of a simpler trained model as attribution scores). Beta-Shapley and LOO are the lowest performing attribution methods whereas TMC-Shapley shows a high performance while still  being worse than ContextCite and Kernel Shap. 

As the sample size increases, all sampling-based methods, except Beta-Shapley show a monotonic improvement, indicating that their approximation does converge towards the true Shapley values with more computational values. Kernel SHAP and ContextCite exhibit the steepest improvement and reach a nearly perfect Pearson correlation (>0.95) and high Kendall's Tau (>0.7 for BIOASK, >0.6 for NQ) once 100 true samples are provided. In contrast, Beta-Shapley shows only marginal improvement or even decrease in performance (precision@3 metric for both NQ and BIOASQ). 

Nevertheless, all methods are effective at identifying the single most important document (or feature) with a precision@1 close to the 0.8 range. Even when k is increasing (k= 3, k= 5), the different methods' performance stay within the same range with negligible changes. 

The performance trends and the relative ranking of the methods stay consistent across both datasets (NQ and BIOASQ), with a slightly higher performance attained in BIOASQ compared to NQ. 

From the obtained results it can be concluded that Kernel SHAP is the most accurate and sample-efficient method for approximating Shapley values with ContextCite following close behind. Both methods offer a substantial improvement over TMC-Shapley, Beta-Shapley and LOO. 

\begin{figure}[H]
    \centering
    \includegraphics[width=0.5\textwidth]{Figures/legend.pdf}

    \begin{subfigure}[b]{0.23\textwidth}
        \includegraphics[width=\textwidth]{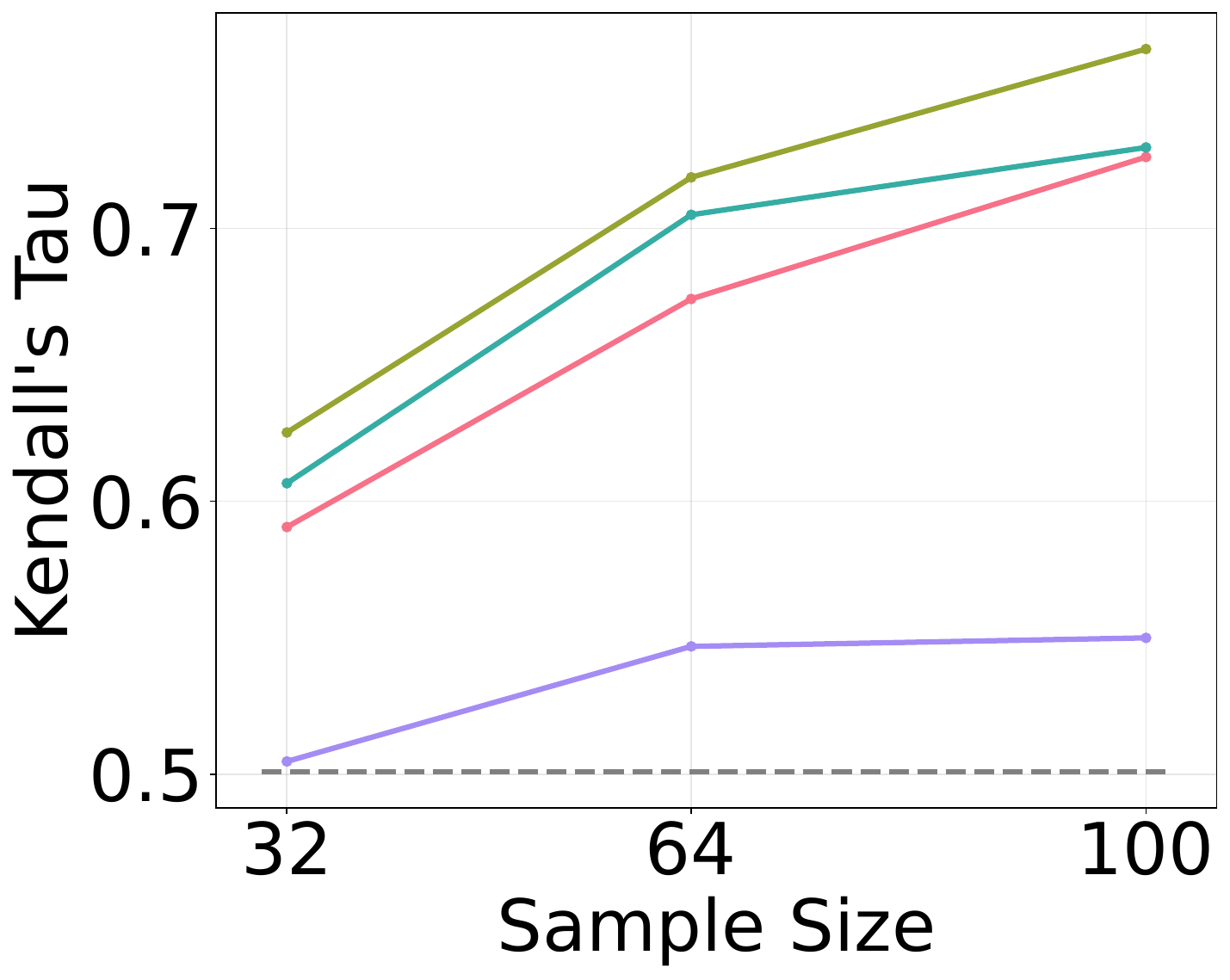}
        \label{fig:bioask_kendall}
    \end{subfigure}
    \hfill
    \begin{subfigure}[b]{0.24\textwidth}
        \includegraphics[width=\textwidth]{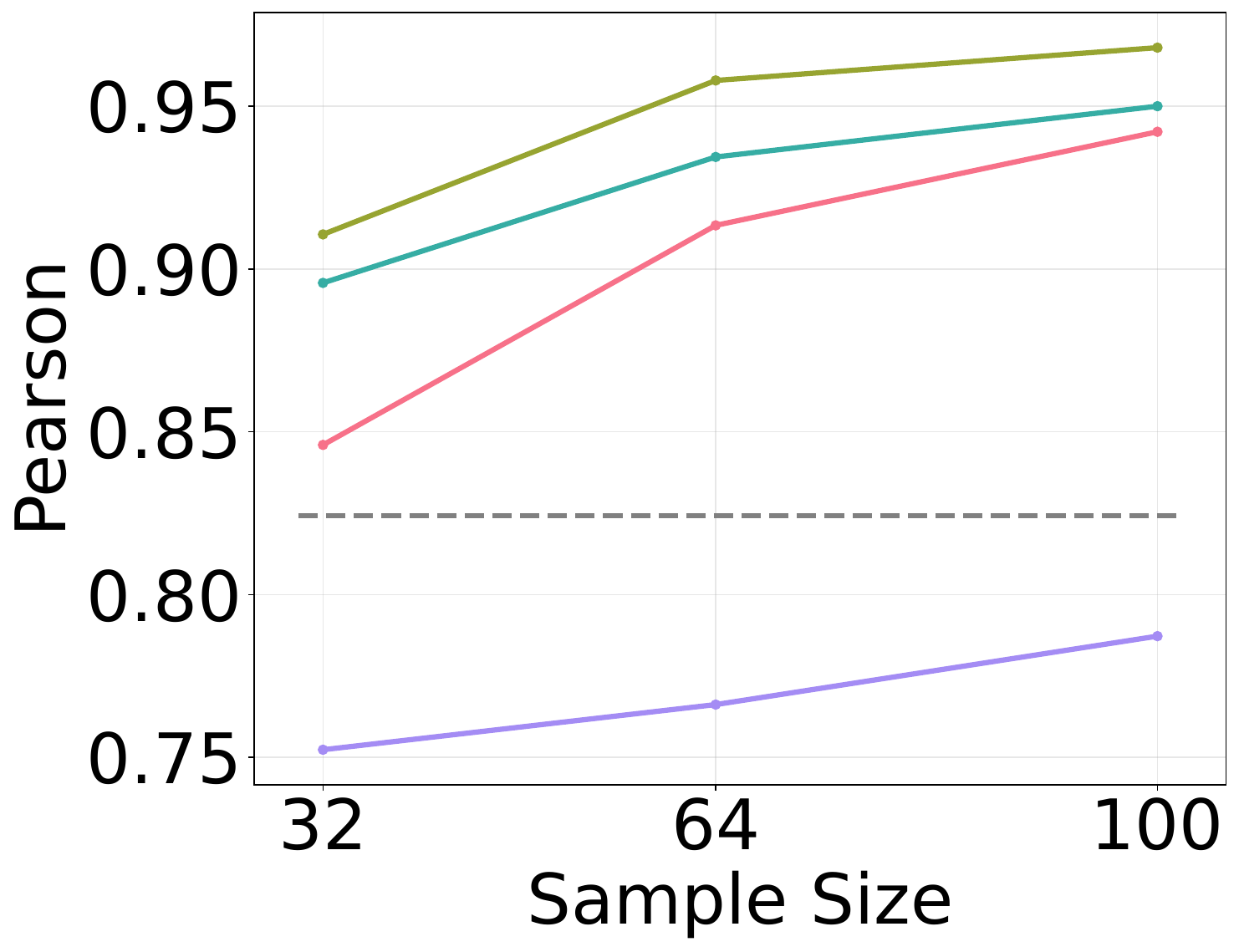}
        \label{fig:bioask_pearson}
    \end{subfigure}
    \hfill
    \begin{subfigure}[b]{0.24\textwidth}
        \includegraphics[width=\textwidth]{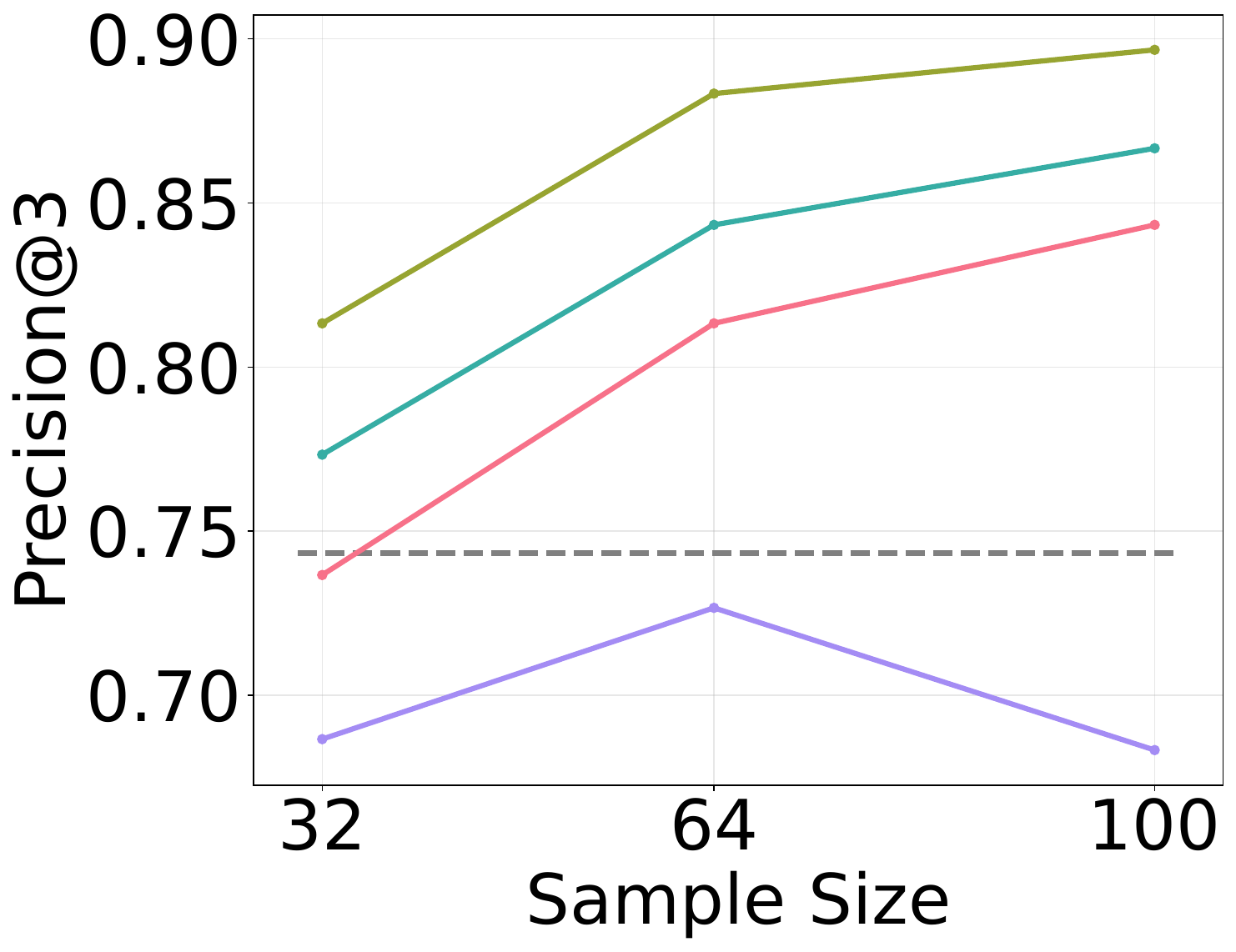}
        \label{fig:bioask_prec3}
    \end{subfigure}
    \hfill
    \begin{subfigure}[b]{0.23\textwidth}
        \includegraphics[width=\textwidth]{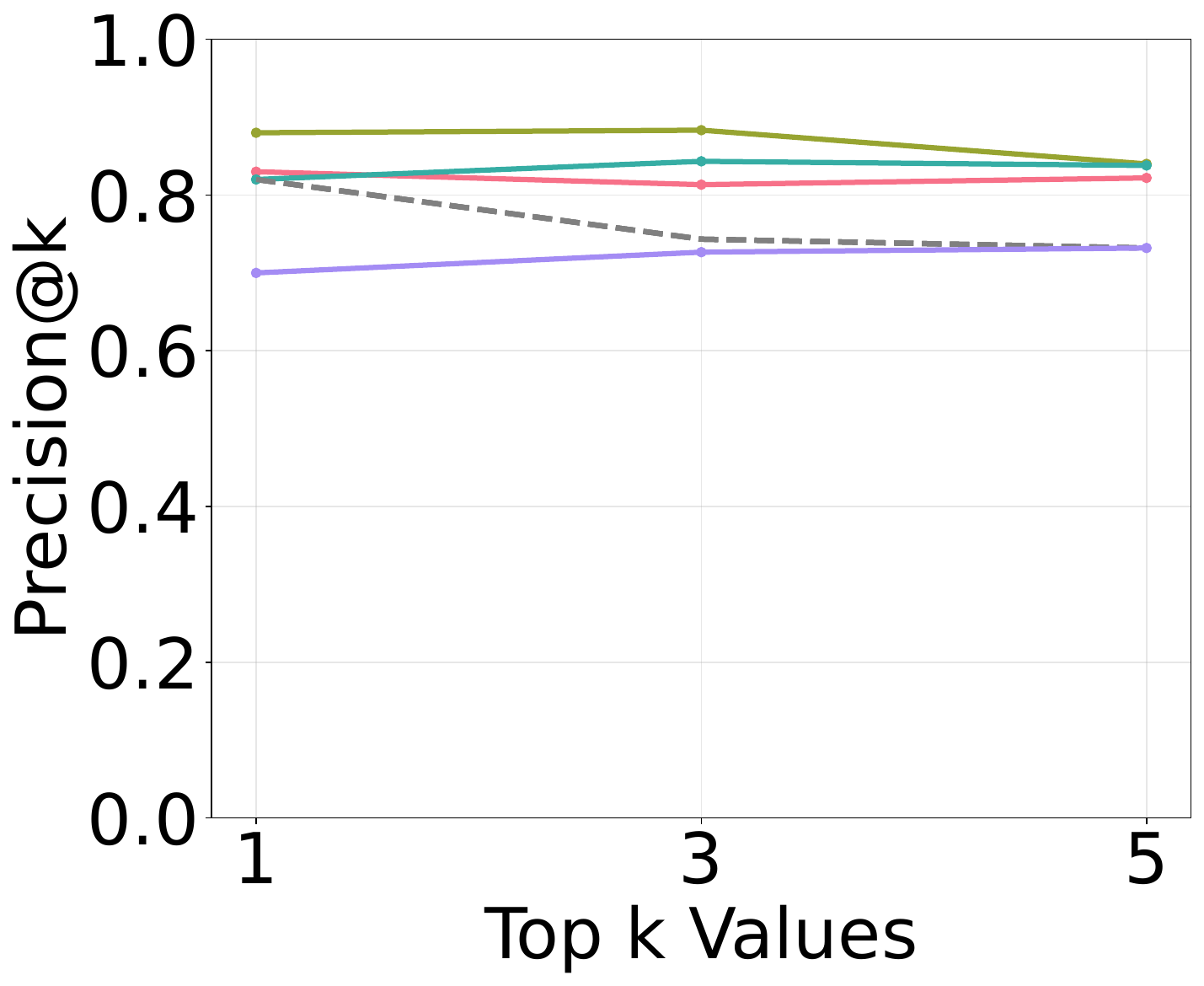}
        \label{fig:bioask_topk}
    \end{subfigure}

    \begin{subfigure}[b]{0.23\textwidth}
        \includegraphics[width=\textwidth]{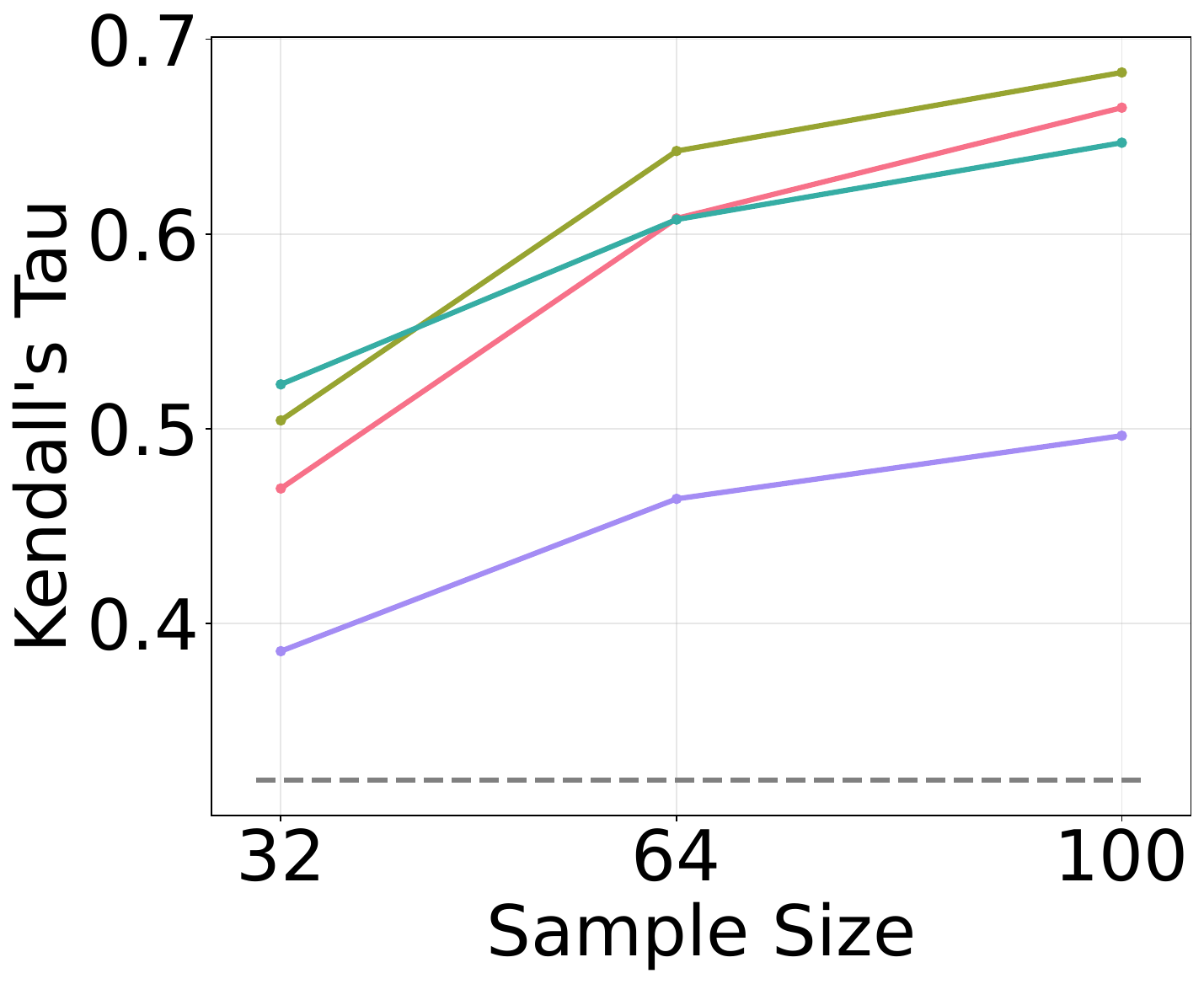}
        \label{fig:nq_kendall}
    \end{subfigure}
    \hfill
    \begin{subfigure}[b]{0.23\textwidth}
        \includegraphics[width=\textwidth]{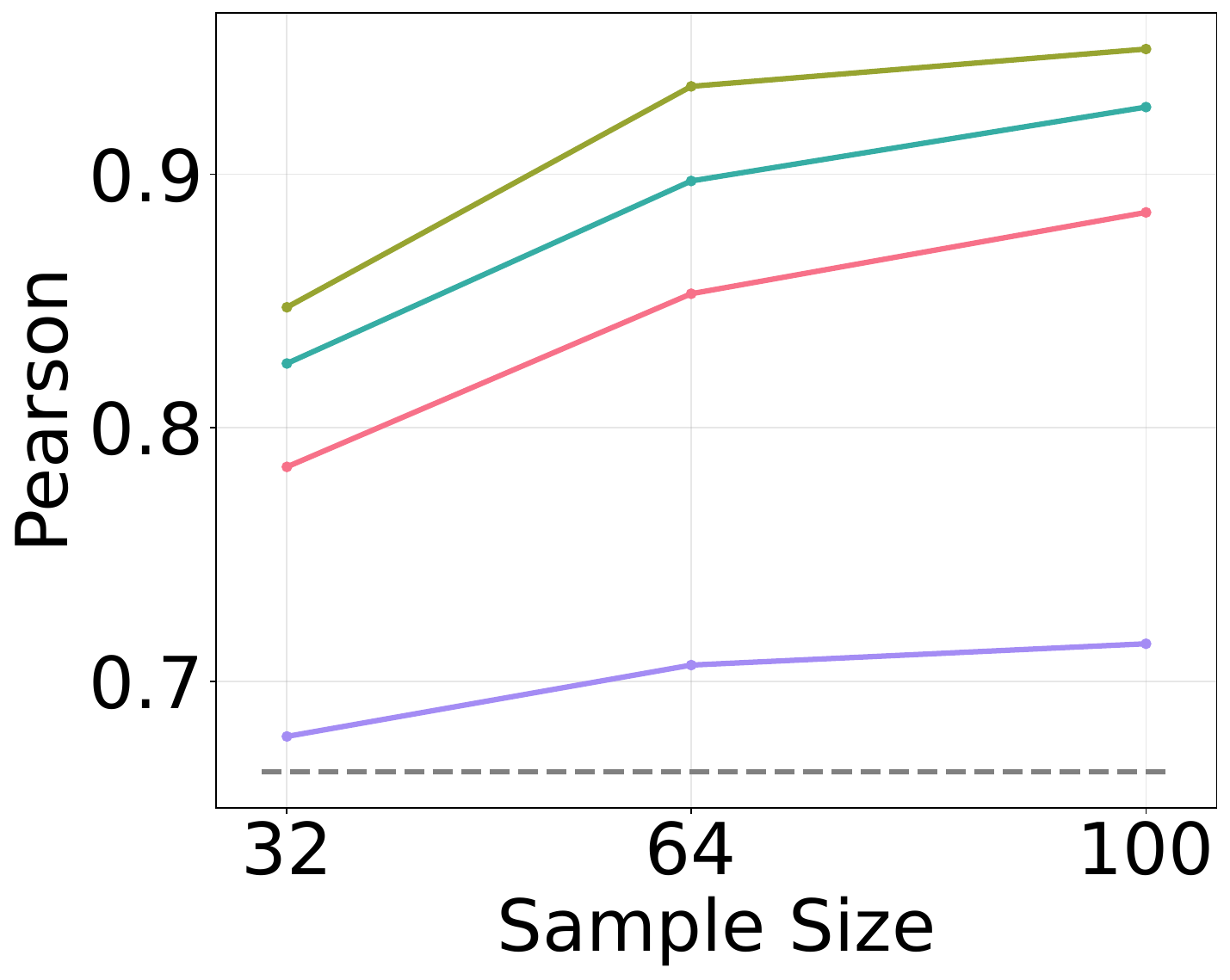}
        \label{fig:nq_pearson}
    \end{subfigure}
    \hfill
    \begin{subfigure}[b]{0.23\textwidth}
        \includegraphics[width=\textwidth]{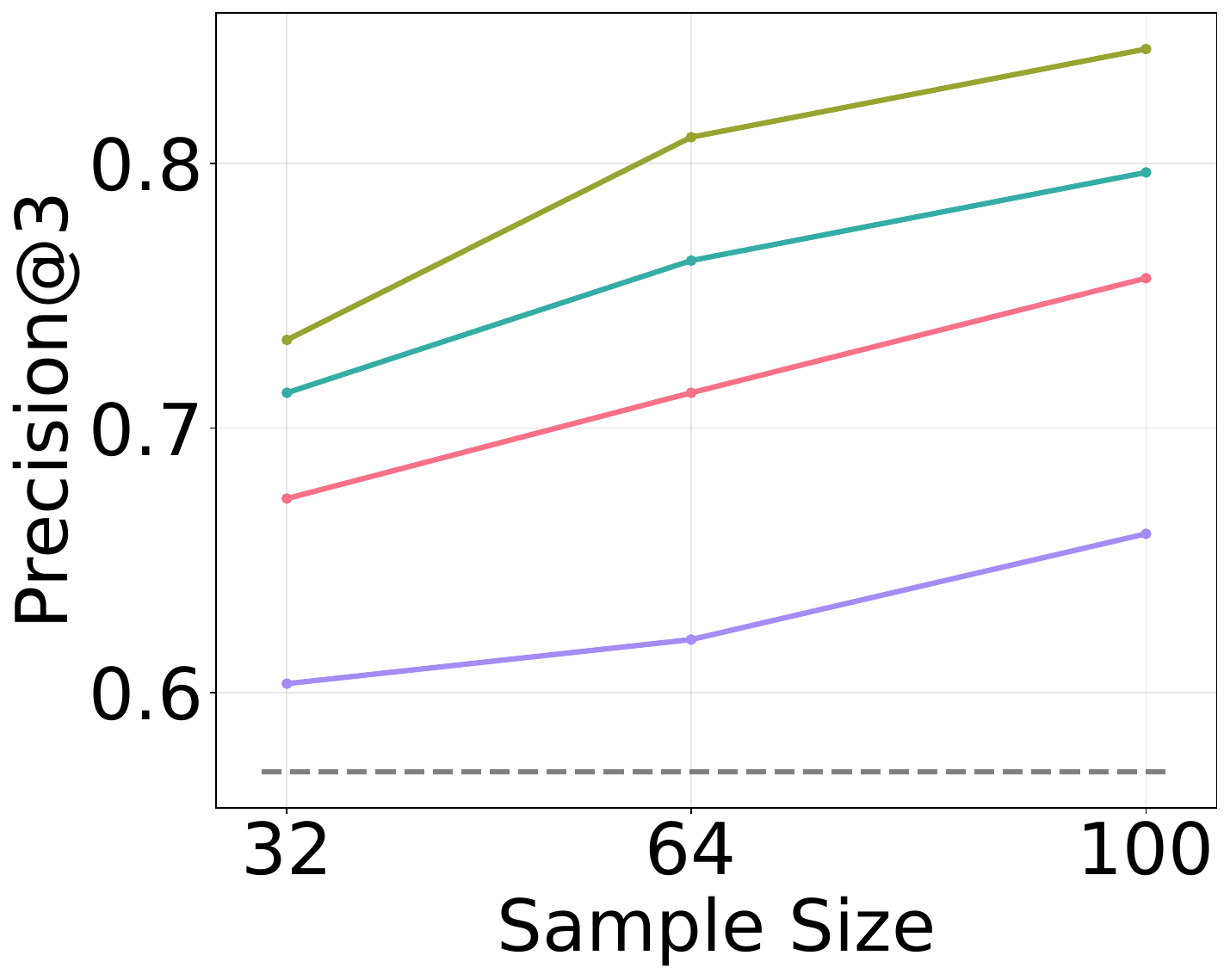}
        \label{fig:nq_prec3}
    \end{subfigure}
    \hfill
    \begin{subfigure}[b]{0.23\textwidth}
        \includegraphics[width=\textwidth]{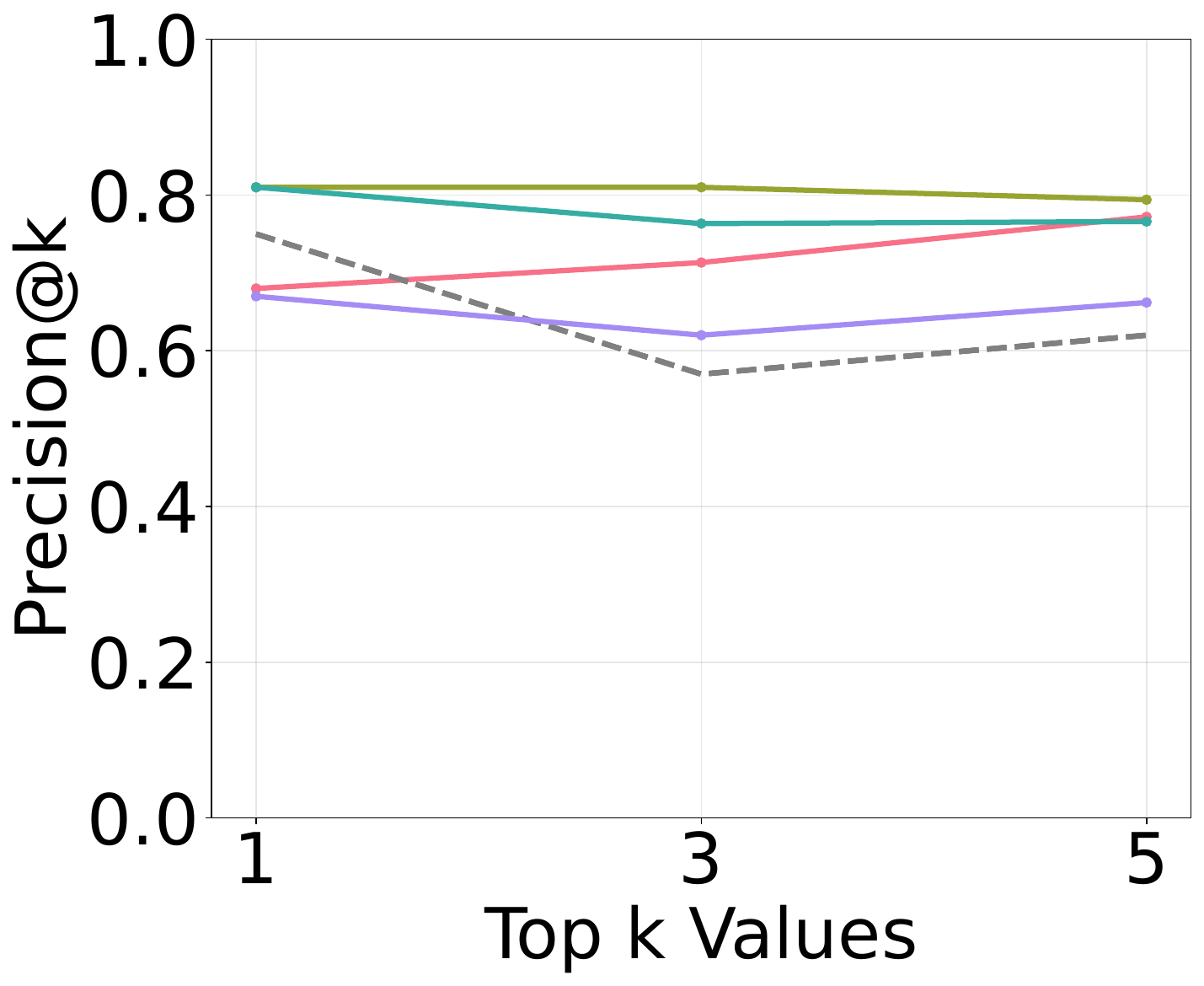}
        \label{fig:nq_topk}
    \end{subfigure}

    \caption{Correlation to Shapley values with Llama-8B using BIOASQ and NQ datasets}
    \label{fig:exp1_llama_all}
\end{figure}

\begin{figure}[H]
    \centering
    \includegraphics[width=0.5\textwidth]{Figures/legend.pdf}

    \begin{subfigure}[b]{0.23\textwidth}
        \includegraphics[width=\textwidth]{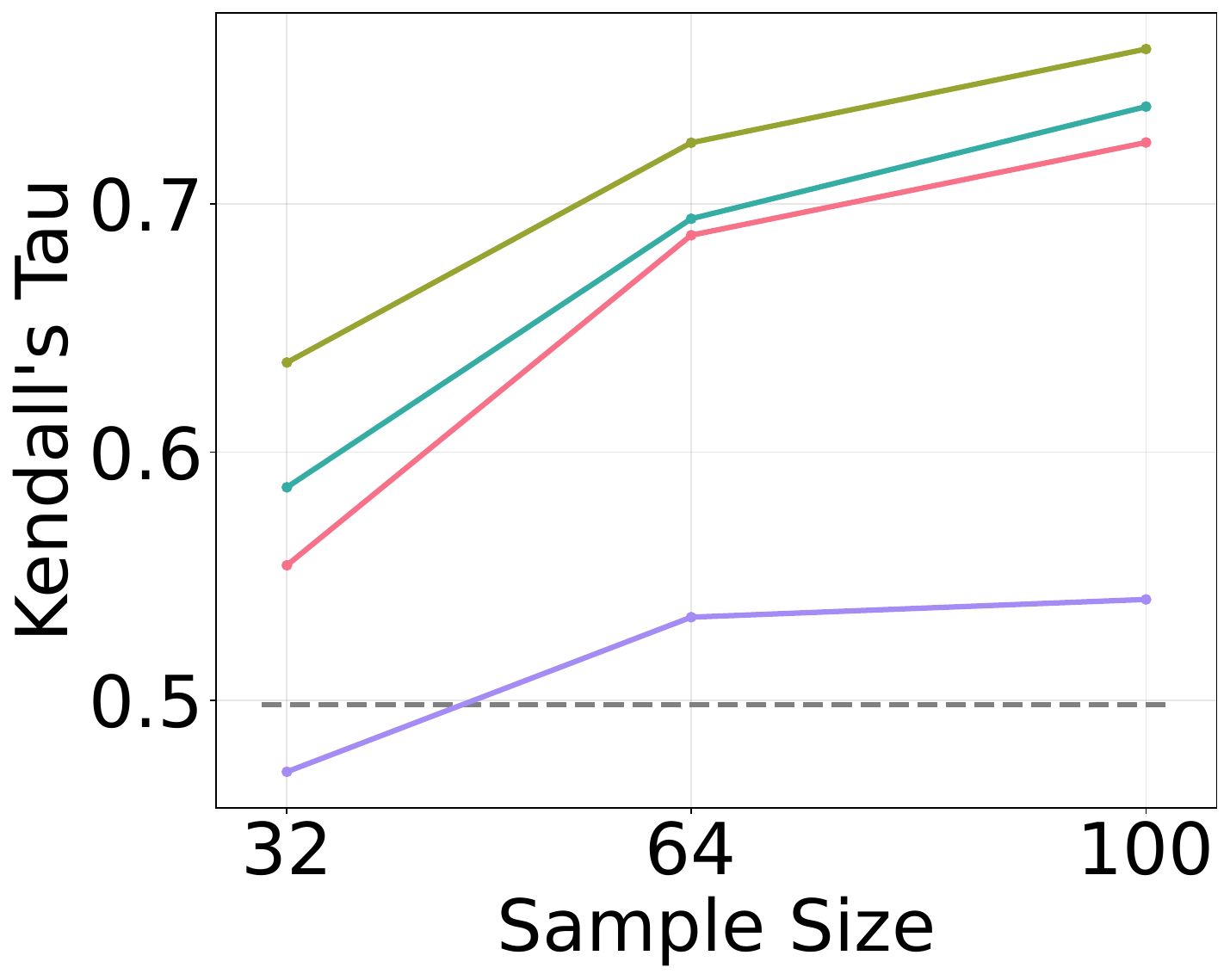}
        \label{fig:bioask_kendall}
    \end{subfigure}
    \hfill
    \begin{subfigure}[b]{0.24\textwidth}
        \includegraphics[width=\textwidth]{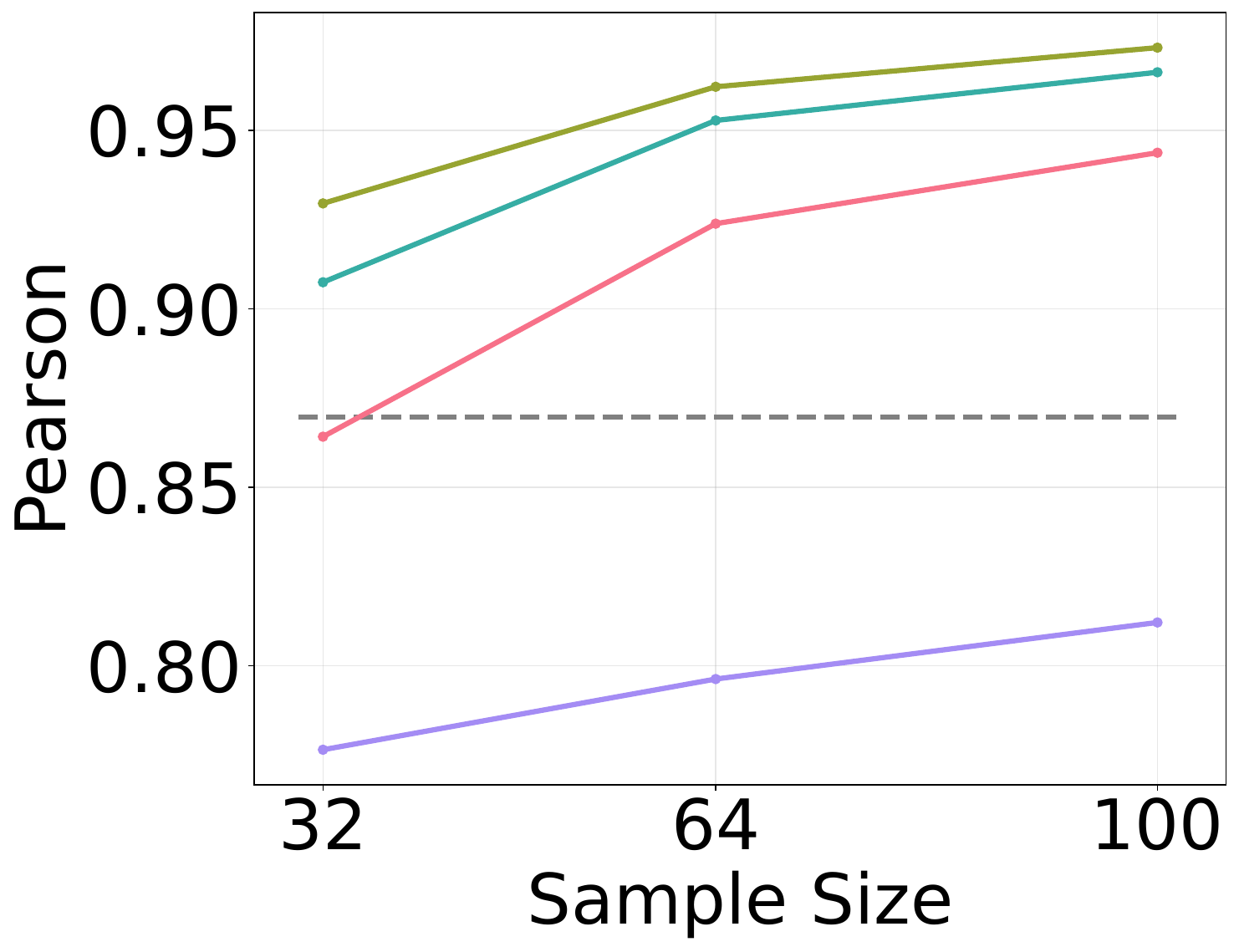}
        \label{fig:bioask_pearson}
    \end{subfigure}
    \hfill
    \begin{subfigure}[b]{0.24\textwidth}
        \includegraphics[width=\textwidth]{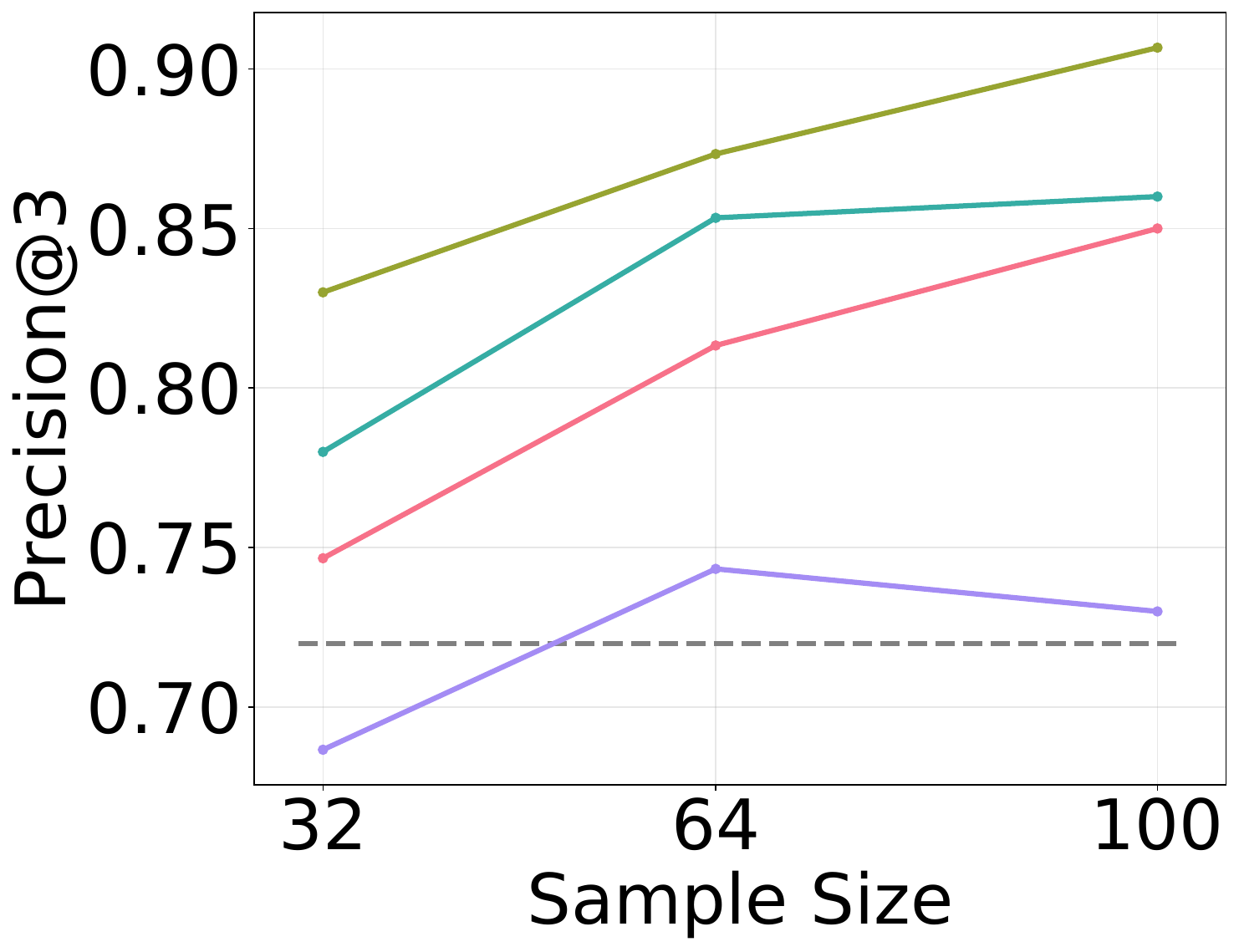}
        \label{fig:bioask_prec3}
    \end{subfigure}
    \hfill
    \begin{subfigure}[b]{0.23\textwidth}
        \includegraphics[width=\textwidth]{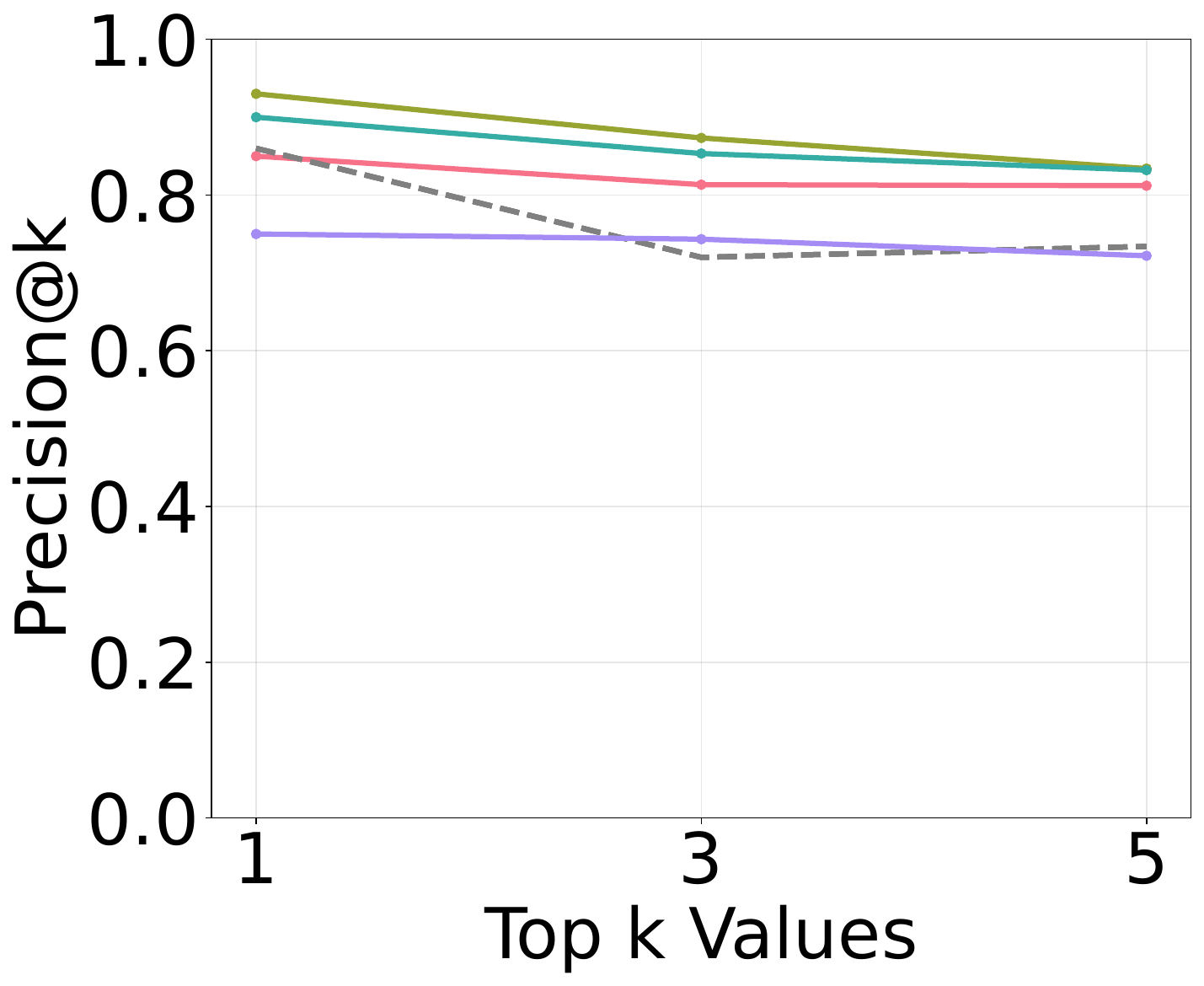}
        \label{fig:bioask_topk}
    \end{subfigure}

    \begin{subfigure}[b]{0.23\textwidth}
        \includegraphics[width=\textwidth]{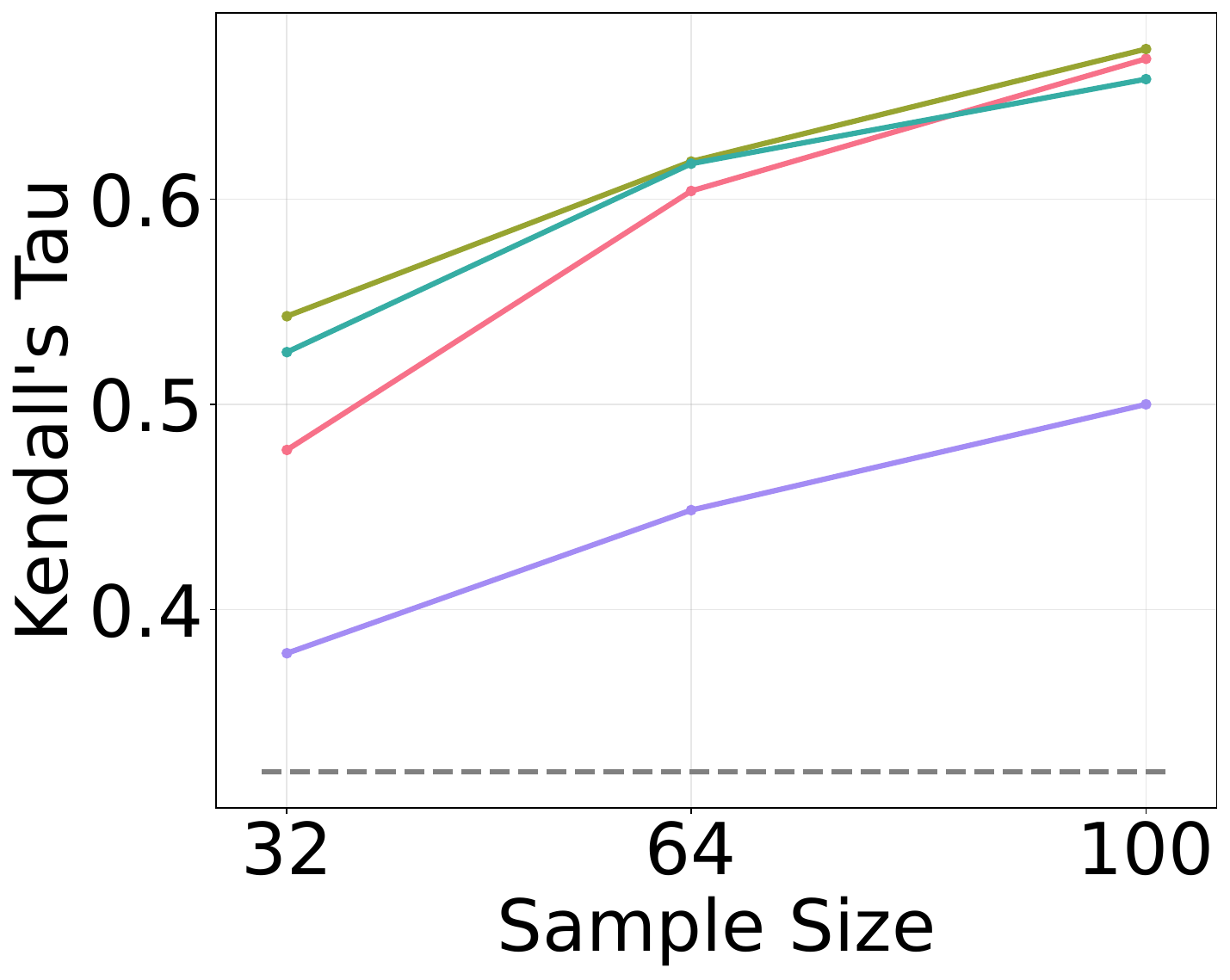}
        \label{fig:nq_kendall}
    \end{subfigure}
    \hfill
    \begin{subfigure}[b]{0.23\textwidth}
        \includegraphics[width=\textwidth]{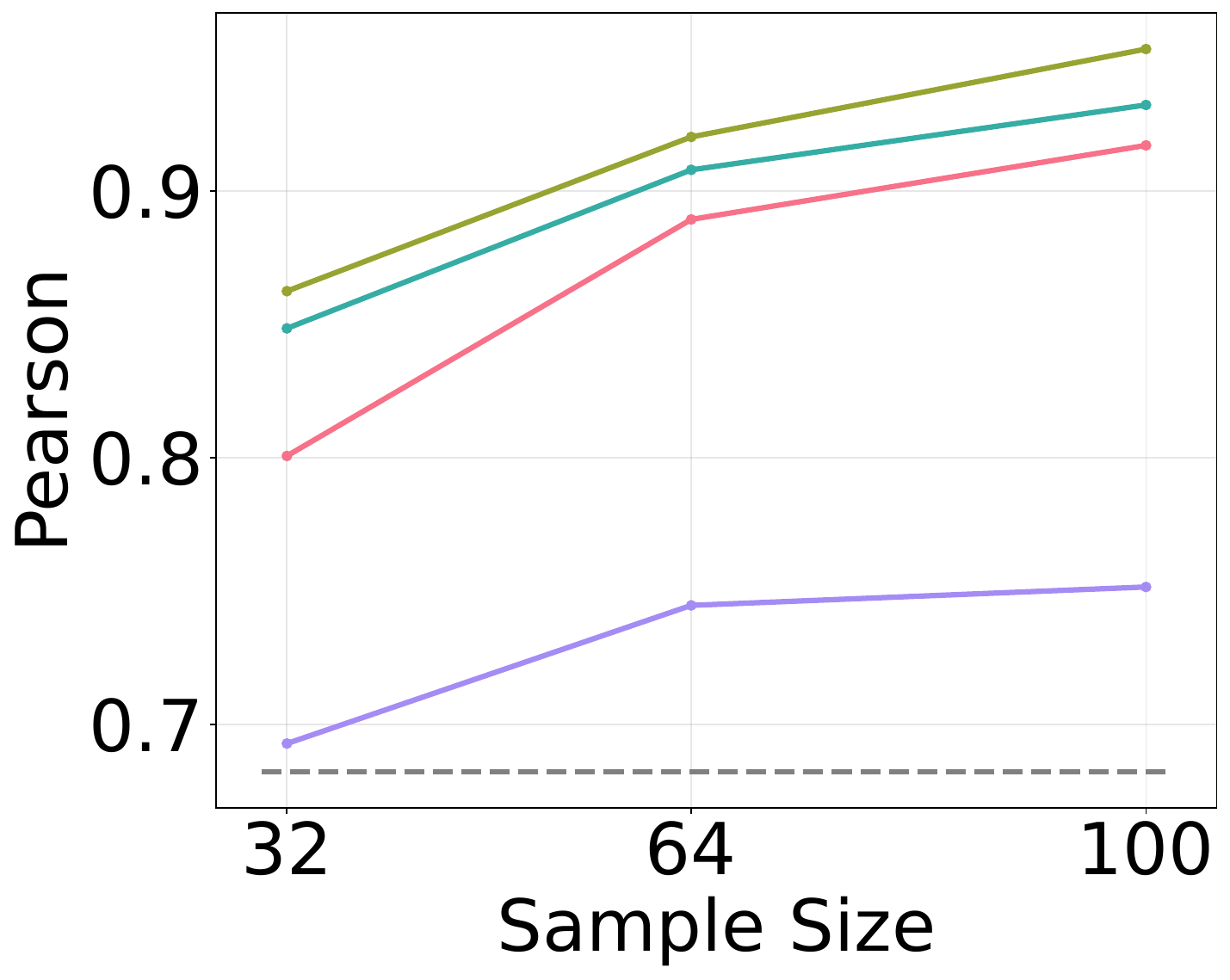}
        \label{fig:nq_pearson}
    \end{subfigure}
    \hfill
    \begin{subfigure}[b]{0.23\textwidth}
        \includegraphics[width=\textwidth]{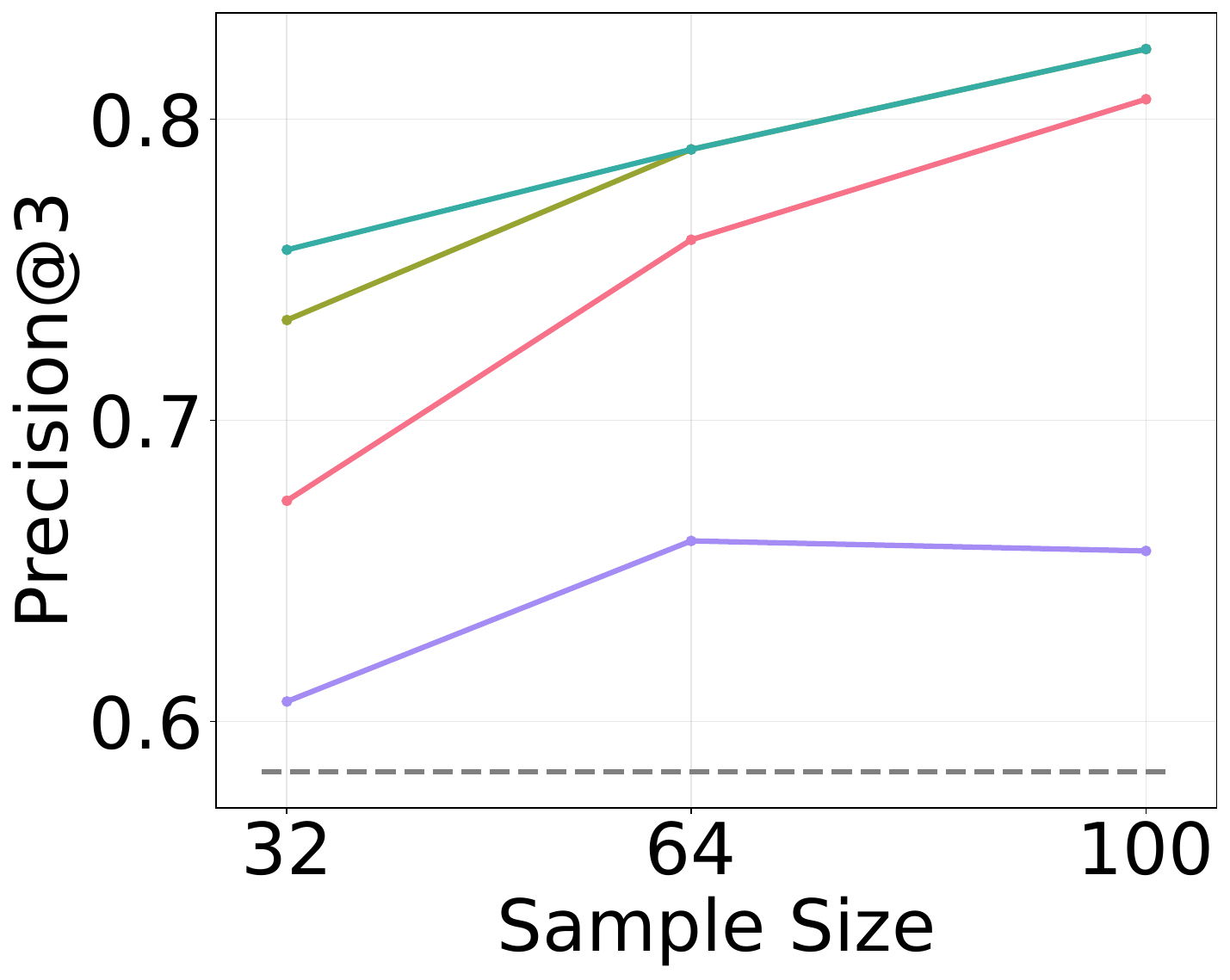}
        \label{fig:nq_prec3}
    \end{subfigure}
    \hfill
    \begin{subfigure}[b]{0.23\textwidth}
        \includegraphics[width=\textwidth]{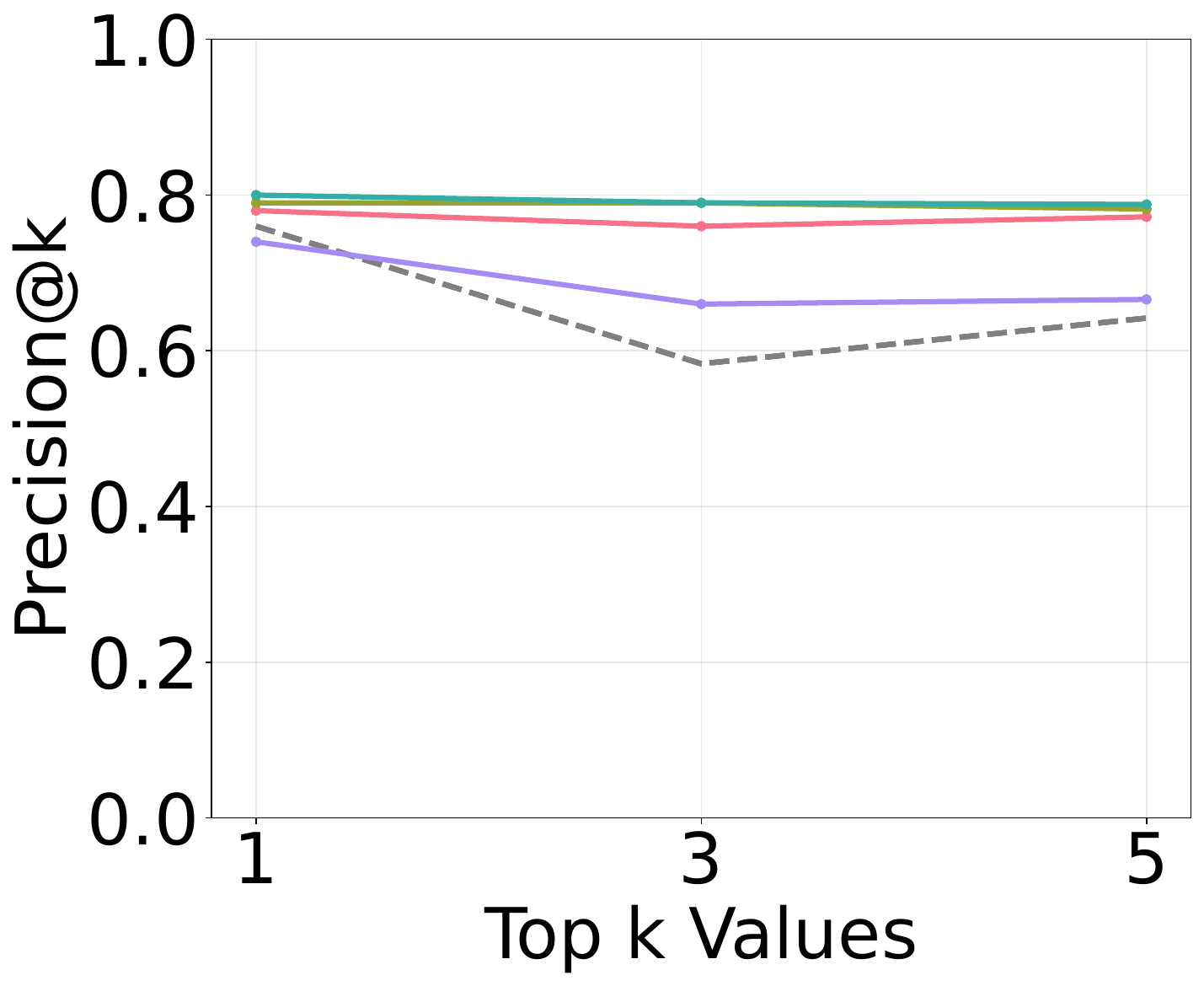}
        \label{fig:nq_topk}
    \end{subfigure}

    \caption{Correlation to Shapley values with Qwen-3B using BIOASQ and NQ datasets}
    \label{fig:exp1_qwen_all}
\end{figure}

\subsection{Experiment 2}
\begin{table}[t]
    \centering
    \scriptsize
    \begin{tabularx}{\textwidth}{bssssssssssss}
        \toprule
         & \multicolumn{4}{l}{\textbf{Qwen 3B}} & \multicolumn{4}{l}{\textbf{Mistral 7B}} & \multicolumn{4}{l}{\textbf{Llama 8B}} \\
        \hline
        $k$ & 2 & 3 & 4 & 5 & 2 & 3 & 4 & 5 & 2 & 3 & 4 & 5 \\
        \midrule
        Shapley & 0.80 & 0.80 & \textbf{0.79} & \textbf{0.78} & \textbf{0.82} & 0.79 & \textbf{0.79} & \textbf{0.80} & 0.80 & 0.79 & \textbf{0.80} & \textbf{0.82} \\ \hline
        TMC-Shapley 32 & 0.72 & 0.69 & 0.70 & 0.70 & 0.75 & 0.67 & 0.67 & 0.70 & 0.69 & 0.71 & 0.72 & 0.71 \\
        Beta Shapley 32 & 0.68 & 0.66 & 0.68 & 0.68 & 0.67 & 0.65 & 0.64 & 0.68 & 0.64 & 0.65 & 0.66 & 0.69 \\
        Kernel SHAP 32 & 0.78 & 0.79 & 0.76 & 0.73 & 0.77 & 0.74 & 0.74 & 0.75 & 0.77 & 0.74 & 0.75 & 0.75 \\
        ContextCite 32 & 0.76 & 0.74 & 0.73 & 0.73 & 0.74 & 0.74 & 0.75 & 0.74 & 0.72 & 0.74 & 0.74 & 0.74 \\ \midrule
        TMC-Shapley 64 & 0.77 & 0.74 & 0.73 & 0.72 & 0.76 & 0.74 & 0.74 & 0.74 & 0.69 & 0.74 & 0.75 & 0.75 \\
        Beta Shapley 64 & 0.67 & 0.69 & 0.68 & 0.67 & 0.71 & 0.67 & 0.65 & 0.69 & 0.62 & 0.66 & 0.67 & 0.69 \\
        Kernel SHAP 64 & 0.80 & 0.79 & 0.78 & 0.76 & 0.80 & 0.79 & 0.77 & 0.77 & 0.79 & 0.78 & 0.79 & 0.80 \\
        ContextCite 64 & 0.79 & 0.79 & 0.77 & 0.76 & 0.78 & 0.79 & 0.76 & 0.75 & 0.76 & 0.79 & 0.80 & 0.79 \\ \midrule
        TMC-Shapley 100 & 0.77 & 0.77 & 0.75 & 0.72 & 0.79 & 0.75 & 0.75 & 0.76 & 0.72 & 0.76 & 0.76 & 0.76 \\
        Beta Shapley 100 & 0.67 & 0.68 & 0.66 & 0.66 & 0.69 & 0.66 & 0.67 & 0.67 & 0.61 & 0.60 & 0.65 & 0.69 \\
        Kernel SHAP 100 & \textbf{0.83} & 0.80 & 0.78 & \textbf{0.78} & \textbf{0.82} & \textbf{0.80} & 0.77 & \textbf{0.80}  & \textbf{0.81} & 0.78 & 0.79 & 0.81 \\
        ContextCite 100 & 0.80 & \textbf{0.81} & 0.78 & 0.77 & 0.80 & 0.79 & 0.77 & 0.77 & 0.78 & \textbf{0.81} & \textbf{0.80} & 0.78 \\ \midrule
        LOO & 0.71 & 0.69 & 0.66 & 0.67 & 0.76 & 0.70 & 0.65 & 0.67 & 0.76 & 0.71 & 0.66 & 0.67 \\
        \bottomrule
    \end{tabularx}
    \caption{Exhaustive top-$k$ for the BioASQ dataset.}
    \label{tab:exhaustive_top_k_bioasq}
\end{table}

Let's discuss other trends that describe the behavior of approximation methods. We can see that larger LLMs tend to yield better peformance. As we can note, across the tables, Mistral-7B-Instruct and LLaMA-3.2-8B-Instruct models tend to achieve higher scores than Qwen-3B-Instruct. This aligns with the general understanding that larger, more capable models can better utilize the provided context to generate more accurate answers. 

Another interesting trend in the data is that as $k$ increases, the performance does not always improve and can sometimes decrease. This phenomenon can be attributed to the ``lost in the middle'' problem, where LLMs can get distracted or confused by an excess of information, some of which might be redundant or conflicting. Thus, a bigger subset of documents does not necessarily provide higher utility to LLMs.



Despite having higher scores for BIOASQ, relative effectiveness of our methods remains stable across both datasets. This suggests that performance of attribution methods is not domain-specific. Their ability to model the marginal contribution of each document is robust, whether the topic is general knowledge or specialized biomedical information.

\end{document}